\documentclass[lettersize,journal]{IEEEtran}
\usepackage{amsmath,amsfonts}
\usepackage{algorithmic}
\usepackage{algorithm}
\usepackage{array}
\usepackage[caption=false,font=normalsize,labelfont=sf,textfont=sf]{subfig}
\usepackage{textcomp}
\usepackage{stfloats}
\usepackage{url}
\usepackage{verbatim}
\usepackage{graphicx}
\usepackage{cite}
\usepackage[table]{xcolor}
\usepackage{booktabs}
\usepackage{multirow}
\usepackage{caption}
\usepackage{hyperref}
\usepackage[switch]{lineno}

\begin{document}

\title{OpenUrban3D: Annotation-Free Open-Vocabulary Semantic Segmentation of Large-Scale Urban Point Clouds}

\author{Chongyu Wang, Kunlei Jing, Jihua Zhu and Di Wang
    \thanks{This work was supported in part by the National Natural Science Foundation of China under Grant 42571444, and in part by the Key Research and Development Program of Shaanxi Province under Grant 2023-YBSF-452. (\textit{Corresponding authors: Di Wang})}
    \thanks{All authors are with the School of Software Engineering, Xi’an Jiaotong University, 710049 Xi’an, China, and also with Shaanxi Joint (Key) Laboratory for Artificial Intelligence (Xi’an Jiaotong University),  Xi'an 710049, China
        (e-mail: chongyu.wang@xjtu.edu.cn, kunlei.jing@xjtu.edu.cn, zhujh@mail.xjtu.edu.cn, diwang@xjtu.edu.cn)}
}



\maketitle

\begin{abstract}
    Open-vocabulary semantic segmentation enables models to recognize and segment objects from arbitrary natural language descriptions, offering the flexibility to handle novel, fine-grained, or functionally defined categories beyond fixed label sets. While this capability is crucial for large-scale urban point clouds that support applications such as digital twins, smart city management, and urban analytics, it remains largely unexplored in this domain. The main obstacles are the frequent absence of high-quality, well-aligned multi-view imagery in large-scale urban point cloud datasets and the poor generalization of existing three-dimensional (3D) segmentation pipelines across diverse urban environments with substantial variation in geometry, scale, and appearance. To address these challenges, we present \textbf{OpenUrban3D}, the first 3D open-vocabulary semantic segmentation framework for large-scale urban scenes that operates without aligned multi-view images, pre-trained point cloud segmentation networks, or manual annotations. Our approach generates robust semantic features directly from raw point clouds through multi-view, multi-granularity rendering, mask-level vision-language feature extraction, and sample-balanced fusion, followed by distillation into a 3D backbone model. This design enables zero-shot segmentation for arbitrary text queries while capturing both semantic richness and geometric priors. Extensive experiments on large-scale urban benchmarks, including SensatUrban and SUM, show that OpenUrban3D achieves significant improvements in both segmentation accuracy and cross-scene generalization over existing methods, demonstrating its potential as a flexible and scalable solution for 3D urban scene understanding.
\end{abstract}

\begin{IEEEkeywords}
    Open-vocabulary semantic segmentation, Large-scale urban point clouds, Annotation-free, Vision-language models, Multi-view projection, Knowledge distillation
\end{IEEEkeywords}

\section{Introduction}
\IEEEPARstart{I}{n} recent years, advances in three-dimensional (3D) point cloud acquisition techniques have transformed large-scale urban 3D mapping, enabling applications in digital twins~\cite{dembski2020urban}, environmental monitoring~\cite{guo2020lidar}, smart cities~\cite{silva2018towards}, urban planning~\cite{urech2020point}, disaster response~\cite{garnett2018lidar}, and autonomous driving~\cite{li2020deep}. Point cloud semantic segmentation, which assigns semantic labels to individual points, serves as a critical bridge from raw geometry to high-level scene understanding that underpins these applications~\cite{guo2020deep}.
\begin{figure}
    \centering
    \includegraphics[width=0.9\linewidth]{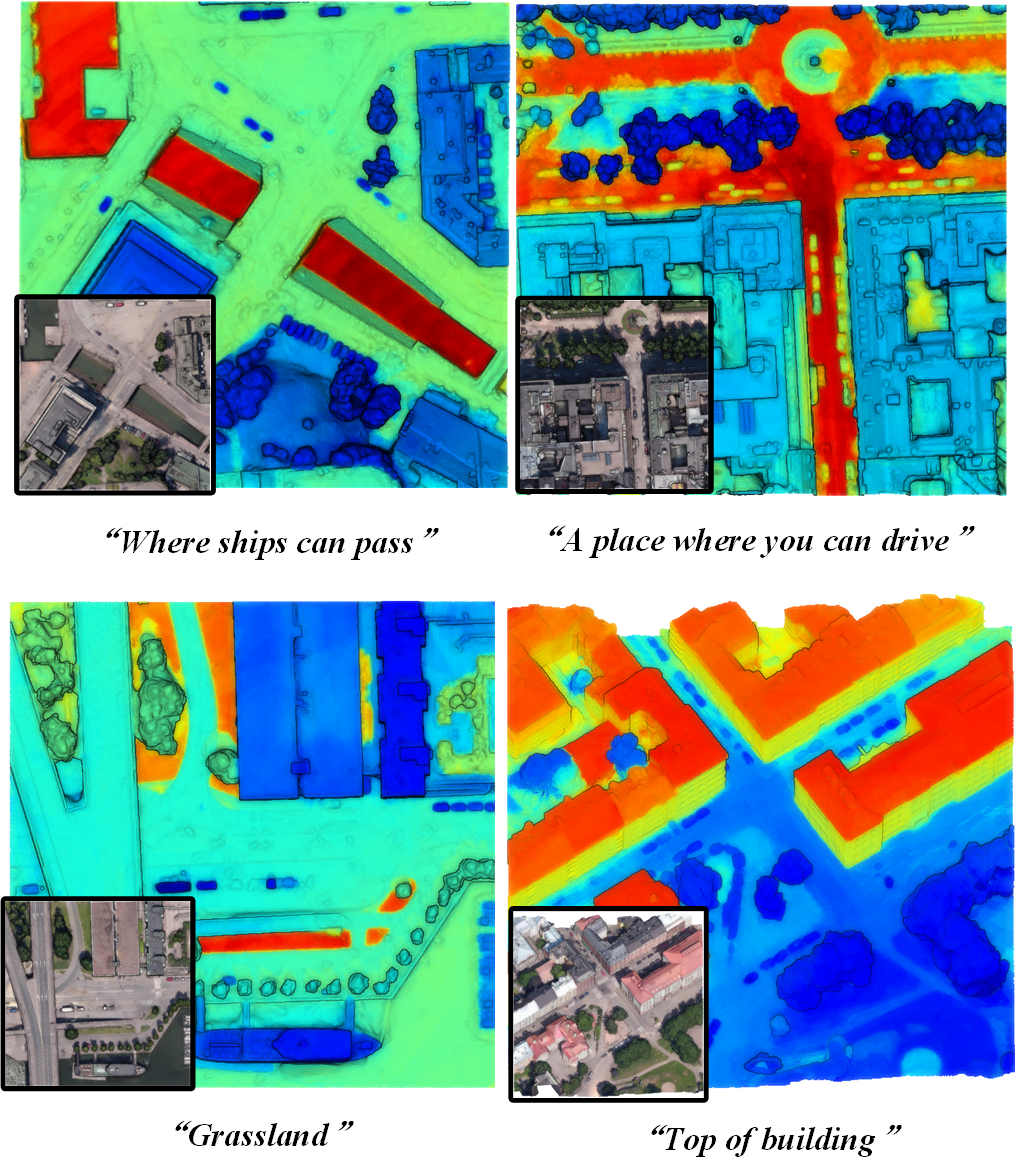}
    \caption{Visualization of open-vocabulary semantic understanding in large-scale urban point clouds using OpenUrban3D. The results are obtained solely from raw point cloud data without manual annotations. Color intensity indicates the similarity to the query text, with dark red denoting high similarity and dark blue denoting low similarity.}
    \label{fig:open_vis_sum}
\end{figure}

Conventional urban point cloud semantic segmentation models rely heavily on supervised training with datasets containing predefined, closed-set categories (e.g., $\sim$20 common object classes), which makes large-scale annotation costly and limits generalization to unseen or fine-grained concepts~\cite{hu2022sqn,sparseconv,pointnet,pointnet++,ptv3,randla}. For instance, a model may identify a ``building'' but fail to distinguish between a ``sidewalk'' and a ``pedestrian plaza'', or to interpret abstract notions such as ``drivable area'' or ``a place for recreation''. These examples highlight the need for more flexible semantic understanding. Such capabilities, as illustrated in Figure~\ref{fig:open_vis_sum}, refer to identifying previously unseen or functionally defined urban elements such as rare infrastructure types, specialized facilities, or context-dependent regions directly from raw point clouds via natural language queries. This flexibility is crucial for applications including rapid disaster assessment, targeted urban planning, and large-scale asset management, where the required semantic categories cannot be exhaustively predefined.

These limitations motivate the exploration of open-vocabulary semantic segmentation~\cite{li2022language,openseg,dong2023maskclip,liang2023open,ODISE}, which enables recognition and segmentation from arbitrary natural language descriptions, rather than being constrained to categories seen during training. While this paradigm has achieved notable success in 2D vision, its extension to large-scale urban point clouds remains underexplored due to two key challenges.
(1) \textit{Lack of high-quality, well-aligned multi-view imagery.} Many successful indoor 3D open-vocabulary methods rely on transferring knowledge from richly annotated 2D images~\cite{peng2023openscene,yang2024regionplc,ding2023pla}, which requires precise spatial alignment between images and point clouds. However, airborne Light Detection and Ranging (LiDAR) and photogrammetry point clouds~\cite{sensaturban,gao2021sum,kolle2021hessigheim}, which constitute the primary data sources for large-scale urban scenes, are often lack of such imagery or provide only low-resolution, single-view colorization images with severe geometric distortions and viewpoint inconsistencies. Furthermore, in many public datasets (e.g., SensatUrban \cite{sensaturban}, SUM \cite{gao2021sum}, H3D \cite{kolle2021hessigheim}, Toronto3D \cite{toronto}), the original images are discarded due to storage constraints, rendering direct 2D-3D feature transfer infeasible.
(2) \textit{Poor generalization of existing segmentation pipelines.} Existing 3D open-vocabulary segmentation models such as OpenMask3D~\cite{takmaz2023openmask3d} and Open3DIS~\cite{nguyen2024open3dis} adopt a ``segment-first, recognize-later'' strategy. In this approach, a pre-trained 3D instance segmentation network first generates class-agnostic masks, and 2D vision models are then used to assign semantic labels to these masks. Although this method is effective in indoor scenes, large-scale urban environments contain vast numbers of object instances with extreme variation in category, geometry, and scale, ranging from small street furniture to entire building complexes. Instance segmentation networks that perform well in controlled indoor datasets (e.g., Mask3D~\cite{schult2022mask3d}) often experience significant performance degradation in such complex settings. In addition, substantial domain gaps arise from differences in city layouts, architectural styles, vegetation types, and sensor configurations, which prevents models trained on one dataset from generalizing to another. As a result, the ``segment-first, recognize-later'' pipeline becomes unreliable because inaccurate class-agnostic masks limit the effectiveness of downstream open-vocabulary recognition.

To address these challenges, we propose \textbf{OpenUrban3D}, an 3D open-vocabulary semantic segmentation framework for large-scale urban scenes. The framework operates without aligned images, pre-trained point cloud segmentation networks, or manual labels. It renders multi-view, multi-granularity projections, extracts mask-level vision-language features, and fuses them via sample-balanced distillation into a 3D backbone. This design enables zero-shot segmentation from arbitrary text queries and paves the way for more flexible and intelligent urban scene understanding.

Overall, our main contributions are as follows:
\begin{itemize}
    \item \textbf{OpenUrban3D framework:} We present the first 3D open-vocabulary semantic segmentation framework for large-scale urban scenes that requires only raw point cloud data as input, eliminating the need for high-quality aligned images or manual annotations.

    \item \textbf{Tailored technical pipeline:} We design a pipeline that integrates multi-view and multi-granularity projection, mask-based feature extraction and unprojection, and 2D-to-3D knowledge distillation. These techniques effectively address the absence of image data and the large variation in object scale.

    \item \textbf{Comprehensive evaluation:} We conduct extensive experiments on two widely used urban scene benchmarks, SensatUrban and SUM, which demonstrate the effectiveness and superiority of the proposed approach.
\end{itemize}

\section{Related Work}

\subsection{Fully Supervised Point Cloud Segmentation}
3D point cloud segmentation assigns a semantic label to each point. Point-based methods such as PointNet~\cite{pointnet} and PointNet++~\cite{pointnet++} learn directly from unordered points via shared Multi-Layer Perceptrons (MLPs) and symmetric pooling, ensuring permutation invariance. Later works such as KPConv~\cite{kpconv} improves local geometry modeling through kernel point convolutions. Voxel-based methods discretize points into regular 3D grids, enabling the use of 2D  convolutional neural network (CNN) architectures. Sparse Convolution~\cite{sparseconv} and MinkowskiNet~\cite{choy20194d} reduce computation by operating only on non-empty voxels. Transformer-based models, including the Point Transformer series~\cite{ptv1,ptv2,ptv3}, employ self-attention to capture long-range dependencies, achieving state-of-the-art results. Despite their accuracy, these methods require large annotated datasets and are restricted to closed-set categories, limiting generalization to novel or rare objects.
\subsection{2D Open-Vocabulary Segmentation}
To overcome closed-set constraints, 2D vision has developed open-vocabulary segmentation using large-scale vision-language models (VLMs)~\cite{clip,alayrac2022flamingo,liu2023visualllava} trained on web-scale image-text pairs.
Pixel-level approaches, such as LSeg~\cite{li2022language}, align per-pixel visual features with text embeddings for arbitrary category recognition. Mask-first approaches, including OpenSeg~\cite{openseg}, Mask-Clip~\cite{dong2023maskclip}, and ODISE~\cite{ODISE}, generate class-agnostic masks and then classify them, often leveraging  the Segment Anything Model (SAM)~\cite{sam} for high-quality, multi-scale segmentation. Our work adopts the mask-first paradigm for its flexibility in large-scale urban scenes.

\subsection{3D Open-Vocabulary Segmentation}
Progress in 3D open-vocabulary segmentation is slower due to scarce 3D-text datasets, leading most methods to use 2D features as an intermediary. Projection-based classification methods, such as PointCLIP~\cite{zhang2022pointclip}, PointCLIPV2~\cite{zhu2023pointclip2}, and CLIP2Point~\cite{huang2023clip2point}, render depth maps and apply 2D encoders for alignment with text, but are limited to classification tasks. Segmentation methods like OpenScene~\cite{peng2023openscene} and CLIP2Scene~\cite{chen2023clip2scene} distill text-aligned 2D features into 3D models leveraging well-aligned depth videos and pre-trained 2D feature extractors; PLA~\cite{ding2023pla} and RegionPLC~\cite{yang2024regionplc} generate text from RGB images to form point cloud–text pairs, but rely on base/novel category partitioning that limits their transferability. OpenGraph~\cite{deng2024opengraph} projects text features from image masks to 3D points. These approaches require well-aligned RGB sequences, often unavailable for large-scale urban data. Mask-based pipelines, such as OpenMask3D~\cite{takmaz2023openmask3d}, Open3DIS~\cite{nguyen2024open3dis}, and OpenIns3D~\cite{huang2024openins3d}, generate class-agnostic masks via 3D instance segmentation models like Mask3D~\cite{schult2022mask3d}, then assign semantics using aligned RGB. However, they struggle with object scale variation, outdoor domain gaps, and occlusions. Suzuki et al.~\cite{suzuki2025open} fuse SAM masks from multi-view projections, but this is impractical for occlusion-heavy urban scenes. Overall, existing methods depend on RGB alignment or pre-trained 3D segmentation backbones, making them unsuitable for raw large-scale urban point clouds.

\begin{figure*}[t]
    \centering
    \includegraphics[height=13cm, keepaspectratio]{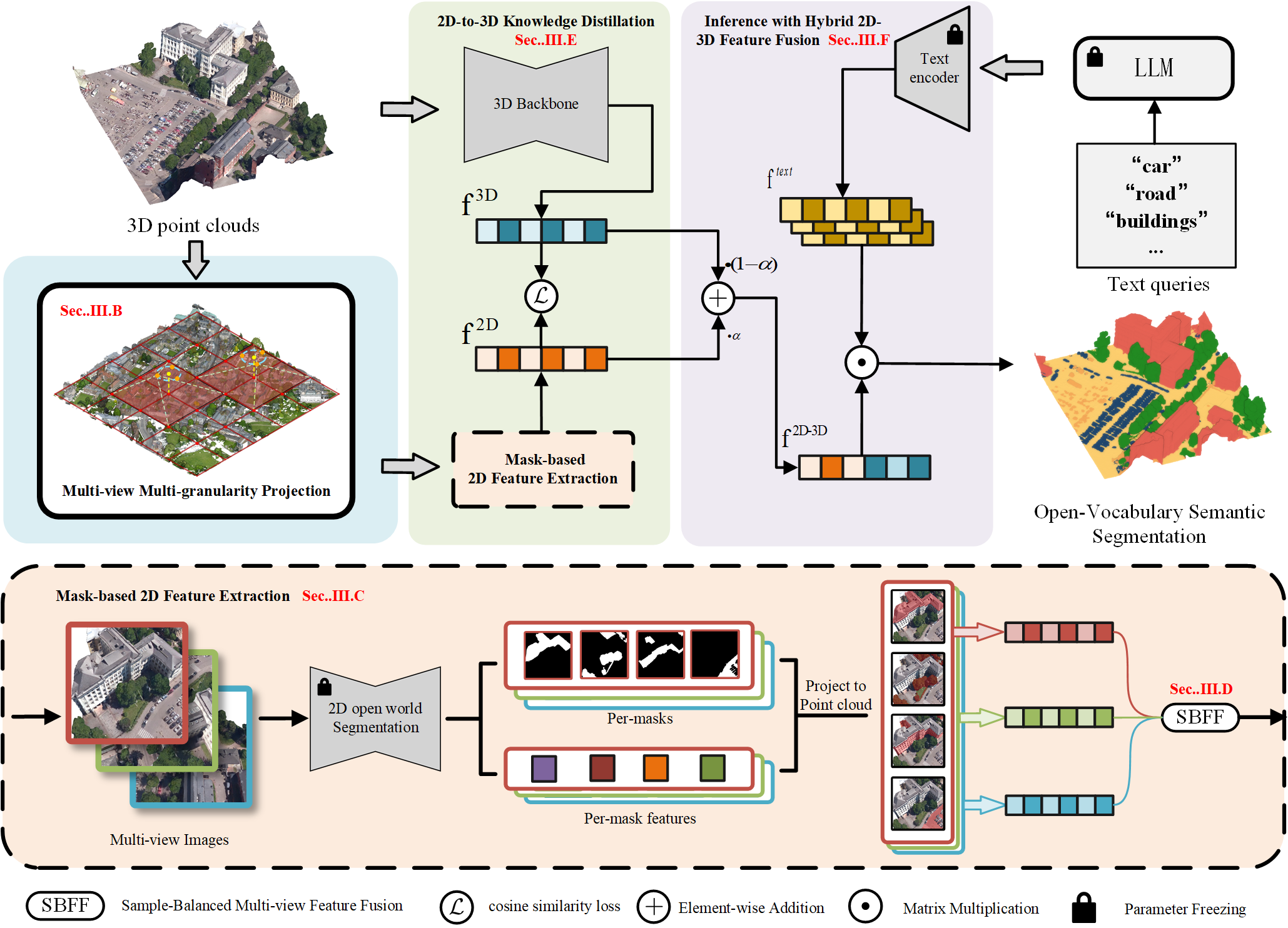}
    \caption{Overall architecture of the proposed OpenUrban3D framework}
    \label{fig:OpenUrban3D_pipline}
\end{figure*}

\section{OpenUrban3D}
The proposed OpenUrban3D is a novel framework enabling 3D open-vocabulary semantic segmentation in large-scale urban scenes. The following describes its overall architecture and key technical modules.

\subsection{Overall Architecture}
The core pipeline of OpenUrban3D, illustrated in Figure~\ref{fig:OpenUrban3D_pipline}, consists of the following steps.

\subsubsection{\textbf{2D Feature Extraction and Construction}}
As highlighted in orange background in Figure~\ref{fig:OpenUrban3D_pipline}, we first generate a series of aligned virtual images for the input point cloud \(P \in \mathbb{R}^{N \times 3}\) using our proposed \emph{Multi-view Multi-granularity Projection} module. A pre-trained 2D vision-language model (VLM), aligned with text features, is then applied to extract per-mask features from these images. The extracted 2D features are mapped back to the original point cloud via unprojection and aggregated using \emph{Sample-balanced Feature Fusion}, resulting in a high-quality 2D feature library $\mathcal{F}_{2D}$ that encodes rich visual semantics for each point.

\subsubsection{\textbf{2D-to-3D Knowledge Distillation}}
As highlighted in green background in Figure~\ref{fig:OpenUrban3D_pipline}, we leverage the geometric structure of the point cloud and mitigate occlusion issues in 2D views through a knowledge distillation strategy. Using $\mathcal{F}_{2D}$ as the teacher signal, we train a 3D backbone so that its per-point 3D features $\mathcal{F}_{3D}$ are semantically aligned with $\mathcal{F}_{2D}$ in the feature space.

\subsubsection{\textbf{Open-Vocabulary Inference}}
As highlighted in purple background in Figure~\ref{fig:OpenUrban3D_pipline}, during the inference stage we fuse the 2D and 3D features to obtain the final per-point representation $\mathbf{F}_{fusion}$. In parallel, a large language model (LLM) processes arbitrary natural language queries to extract the corresponding target class descriptions. The cosine similarity between $\mathbf{F}_{fusion}$ and the target text embeddings is then computed to produce the open-vocabulary semantic segmentation of the point cloud.

\begin{figure*}[t]
    \centering
    \includegraphics[height=7cm, keepaspectratio]{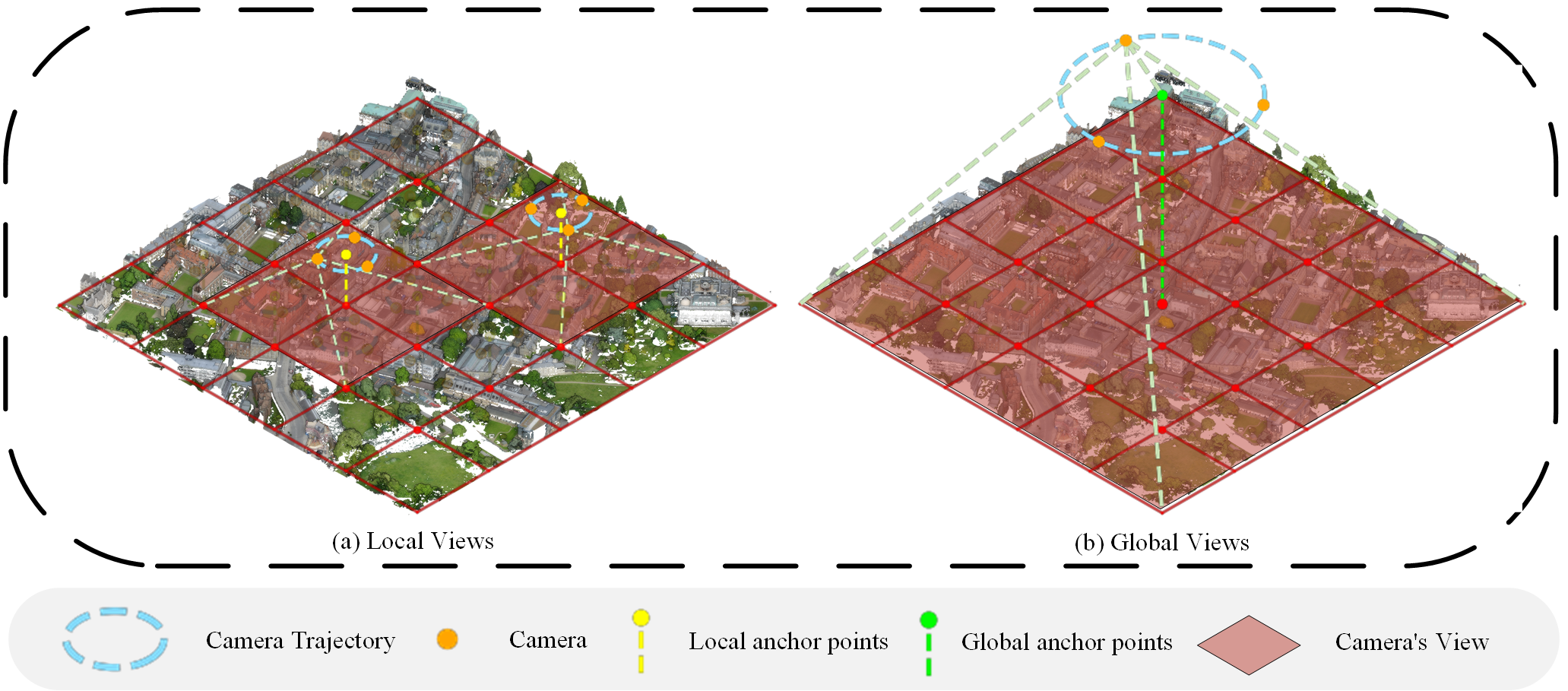}
    \caption{Overview of the Multi-view Multi-granularity Projection module.}
    \label{fig:MMP}
\end{figure*}

\subsection{Multi-view Multi-granularity Projection}
\label{sec:MMGP}
To address the challenges of missing imagery and large variations in object scale in large-scale urban scenes, we design a projection module that generates information-rich rendered images, as illustrated in Figure~\ref{fig:MMP}.

\subsubsection{Camera Pose Generation}
We define global and local anchor points within the scene to guide the viewpoints of the virtual camera, aiming to capture both large-scale and small-scale objects, respectively. Before computing the poses, we first determine the bounding box of the point cloud scene, denoting its dimensions as length $L$, width $W$, and height $H$, with the origin of its coordinate system at $(x_0, y_0, z_0)$.

\paragraph{Global Views}
As shown in the right part of Figure~\ref{fig:MMP}, a global anchor point is established directly above the center of the point cloud’s bounding box, with coordinates
\begin{equation}
    \label{eq:global_anchor}
    A_g = \left(c_x,\, c_y,\, H + \sqrt{L \cdot W}\right),
\end{equation}
where
\begin{equation}
    \label{eq:global_center}
    (c_x,\, c_y,\, c_z) = \left(x_0 + \tfrac{W}{2},\; y_0 + \tfrac{L}{2},\; z_0 + \tfrac{H}{2}\right)
\end{equation}
represents the center of the bounding box. The camera moves along a horizontal circular path centered at this anchor point with a radius
\begin{equation}
    \label{eq:global_radius}
    r_g = \tfrac{\sqrt{L \cdot W}}{4}
\end{equation}
and always points towards the target point
\begin{equation}
    \label{eq:global_target}
    T_g = \left(c_x,\, c_y,\, \tfrac{1}{2}\left(H + \sqrt{L \cdot W}\right)\right).
\end{equation}
The world coordinates of the $k$-th global camera on the circular path are
\begin{equation}
    \label{eq:global_cam_coords}
    \mathbf{C}^k =
    \begin{cases}
        x_k = c_x + r_g \cdot \cos(\theta_k), \\
        y_k = c_y + r_g \cdot \sin(\theta_k), \\
        z_k = H + \sqrt{L \cdot W},
    \end{cases}
\end{equation}
where the viewing angle $\theta_k$ is defined as
\begin{equation}
    \label{eq:theta}
    \theta_k =
    \begin{cases}
        \theta_k = \theta_0 + kA,        \\
        \theta_0 \sim \mathcal{U}(0, A), \\
        k = 0, 1, \dots, \frac{360}{A} - 1,
    \end{cases}
\end{equation}
with $A$ denoting the fixed angular interval and $\theta_0$ the random initial angular offset sampled uniformly from $(0, A)$. The global camera pose is then computed from its world coordinates $A_g$ and the target point $T_g$ using a standard \texttt{lookat} function.

\paragraph{Local Views}
As shown in the left part of Figure~\ref{fig:MMP}, to capture fine-grained local details we introduce local anchor points controlled by a hyperparameter $K$. Specifically, we construct a grid on the scene's horizontal plane (XY-plane) composed of $K+1$ equidistant vertical lines and $K+1$ equidistant horizontal lines. The $(K+1) \times (K+1)$ intersection points of this grid are defined as the horizontal coordinates for the local anchor points.

The coordinates of the $(i,j)$-th local anchor point $\mathbf{a}_{ij}$ are given by
\begin{equation}
    \label{eq:local_anchor}
    \mathbf{a}_{ij} = \left( x_0 + i \cdot \frac{W}{K+1},\; y_0 + j \cdot \frac{L}{K+1},\; H + \frac{\sqrt{L \cdot W}}{2K} \right),
\end{equation}
where $i, j \in \{1, \dots, K\}$.

The camera orbits each local anchor point $\mathbf{a}_{ij}$ in a horizontal circular path with radius
\begin{equation}
    \label{eq:local_radius}
    r_l=\tfrac{\sqrt{L \cdot W}}{R\cdot K}
\end{equation}
where $R$ is a hyperparameter controlling the radius size. The world coordinates of the camera orbiting the $(i,j)$-th local anchor point are
\begin{equation}
    \label{eq:local_cam_coords}
    \mathbf{c}_{ij}^k =
    \begin{cases}
        x_{ij}^k = a_{ij,x} + r_l \cdot \cos(\theta_k), \\
        y_{ij}^k = a_{ij,y} + r_l \cdot \sin(\theta_k), \\
        z_{ij}^k = a_{ij,z},
    \end{cases}
\end{equation}
where $\theta_k$ is defined in Equation~\ref{eq:theta}.

Throughout this motion, the camera is oriented towards the target point
\begin{equation}
    \label{eq:local_target}
    \mathbf{t}_{ij} = \left( x_0 + i \cdot \tfrac{W}{K+1},\; y_0 + j \cdot \tfrac{L}{K+1},\; H + \tfrac{\sqrt{L \cdot W}}{2} \right),
\end{equation}
ensuring that the line of sight remains focused on the center of the local region. The camera pose is then computed from its world coordinates $\mathbf{a}_{ij}$ and the target point $\mathbf{t}_{ij}$ using a standard \texttt{lookat} function.

\subsubsection{Generation of Camera Intrinsics}
After determining the camera extrinsic parameters (position and pose), we define the intrinsic parameters for each camera type. The key step is to compute an appropriate Field of View (FoV) that ensures precise coverage of the intended scene region.

\paragraph{Global Cameras}
For each camera orbiting the global anchor point
$(c_x,\, c_y,\, H + \tfrac{\sqrt{L \cdot W}}{2})$,
the intrinsics are configured based on a virtual camera placed at the anchor and oriented vertically downwards. The FoV is set to fully encompass the entire scene bounding box of dimensions $L \times W$.

\paragraph{Local Cameras}
For each camera orbiting a local anchor point $\mathbf{a}_{ij}$, the intrinsics are defined in the same way. A virtual camera is placed at the local anchor, looking vertically downwards, and its FoV is set to cover exactly the local grid area of size $\tfrac{W}{K} \times \tfrac{L}{K}$ centered at that anchor.

After processing through the Multi-view Multi-granularity Projection module, the input point cloud
$P \in \mathbb{R}^{N \times 3}$
is projected into a set of rendered images:
\[
    \mathcal{V} = \{ v_1, v_2, \dots, v_m \}, \quad v_i \in \mathbb{R}^{h \times w \times 3}.
\]

\subsection{Mask-based 2D Feature Extraction and Back-Projection}
As highlighted in orange background in Figure~\ref{fig:OpenUrban3D_pipline}, we propose a mask-based feature extraction and back-projection module that aligns 2D semantic features with 3D points. Object-level mask features are extracted using a pre-trained vision-language model and are then accurately projected onto the point cloud through camera geometry and depth validation.

\subsubsection{Feature Extraction}
To address the challenges posed by large variations in object scale and the lower quality of rendered images compared to real photographs in large-scale urban scenes, we adopt an object-level feature generation strategy. Specifically, we employ a pre-trained and frozen 2D vision-language model capable of mask-based feature extraction. This choice is motivated by the fact that object-level mask features are more robust to ambiguous boundaries and scale variations than pixel-level features, and are better suited for the subsequent sample-balanced multi-view feature fusion.

Given an input rendered RGB image $\mathbf{v}_i \in \mathbb{R}^{h \times w \times 3}$, the model outputs $K$ binary masks $\{\mathbf{M}_k \in \{0,1\}^{h \times w}\}_{k=1}^K$ together with their corresponding feature vectors $\{\mathbf{f}_k \in \mathbb{R}^{C}\}_{k=1}^K$, where $C$ denotes the feature dimension.

\subsubsection{Feature Back-Projection}
Once the 2D features have been extracted, the next step is to associate them precisely with the 3D point cloud. For any point $\mathbf{p} \in \mathbb{R}^3$ in the scene’s point cloud $P \in \mathbb{R}^{N \times 3}$, the back-projection proceeds as follows. For each rendered image $v_i$, the point $\mathbf{p}$ is projected onto its pixel plane using the corresponding camera intrinsic matrix $\mathbf{I}_i$ and extrinsic matrix $\mathbf{E}_i$, following the standard pinhole camera model:
\begin{equation}
    \label{eq2}
    \tilde{\mathbf{u}} = \mathbf{I}_i \mathbf{E}_i \tilde{\mathbf{p}},
\end{equation}
where $\tilde{\mathbf{p}}$ and $\tilde{\mathbf{u}}$ are the homogeneous coordinates of the 3D point $\mathbf{p}$ and its corresponding 2D pixel coordinate $\mathbf{u} = (u,v)$, respectively.

To handle occlusions, we use the depth map (Z-buffer) generated concurrently during rendering. A projection is considered valid only if the depth value of $\mathbf{p}$ matches the value stored in the depth map at pixel $\mathbf{u}$, ensuring that only the points closest to the camera are assigned features.

For a valid projection, we determine the mask index $k$ such that $\mathbf{M}_k(\mathbf{u}) = 1$. The corresponding feature vector $\mathbf{f}_k$ is then assigned to the 3D point $\mathbf{p}$ as its 2D visual representation from the rendered image $v_i$.

\subsection{Sample-Balanced Multi-view Feature Fusion}
A single 3D point may be projected into multiple views, necessitating the fusion of features from these different perspectives. However, the vast disparity in the quantity and scale of objects within urban scenes (e.g., ``buildings'' versus ``vehicles'') leads to severe data imbalance. During knowledge distillation, this imbalance can bias the model towards large, over-represented objects, thereby impairing its ability to learn features for smaller objects. To address this, we propose the following sample-balanced feature fusion strategy:

\begin{enumerate}
    \item \textbf{Identifying Over-sampled Masks:}
          For each view, count the number of projected points in each mask. Compute the threshold $\tau$ as the average number of points in the top-$k$ masks with the fewest points:
          \begin{equation}
              n_{v,j} = \big|\{\mathbf{p}_i \in \mathcal{P} \mid \mathbf{p}_i \mapsto M_j \}\big|,
          \end{equation}
          \begin{equation}
              \tau = \frac{1}{k} \sum_{j \in \mathcal{I}_k} n_{v,j},
          \end{equation}
          where $\mathcal{I}_k$ denotes the indices of the top-$k$ masks with the fewest points.

    \item \textbf{Balanced Sampling:}
          For any ``large'' mask $M_j$ with $n_{v,j} > \tau$, randomly down-sample to a quantity comparable to $\tau$:
          \begin{equation}
              \tilde{M}_j =
              \begin{cases}
                  \text{RandSample}(M_j, \tau), & n_{v,j} > \tau,   \\
                  M_j,                          & n_{v,j} \le \tau.
              \end{cases}
          \end{equation}

    \item \textbf{Feature Fusion:}
          Fuse the multi-view features only from views where the point $\mathbf{p}$ is retained after sampling. The initial 2D feature $\mathbf{f}_{2D,p}$ is obtained by averaging over these valid views:
          \begin{equation}
              \mathcal{V}_p = \{v \mid \mathbf{p} \in \tilde{M}_j \text{ in view } v\},
              \quad
              \mathbf{f}_{2D,p} = \frac{1}{|\mathcal{V}_p|} \sum_{v \in \mathcal{V}_p} \mathbf{f}_{v,p}.
          \end{equation}
\end{enumerate}

Through this procedure, we construct a balanced 2D feature library
\(
F_{2D} = \{ f_{2D,p} \mid p \in P_r \},
\)
where \(P_r \subseteq P\) denotes the subset of points that can be rendered, effectively mitigating the data imbalance problem.

\subsection{2D-to-3D Knowledge Distillation}
As highlighted in green background in Figure~\ref{fig:OpenUrban3D_pipline}, although 2D features $\mathcal{F}_{2D}$ derived from rendered images provide rich semantic information, they have inherent limitations. First, due to the complexity of urban environments, such as mutual occlusions between high-rise buildings, the rendered views cannot achieve complete coverage of all 3D points. Second, relying solely on 2D projections introduces viewpoint inconsistencies and fails to fully exploit the valuable 3D geometric information inherent in the point cloud.

To overcome these issues and enable the model to perceive 3D geometric structures, we adopt a knowledge distillation strategy, following the approach of OpenScene \cite{peng2023openscene}. This involves transferring the 2D vision-language knowledge to a 3D encoder $\mathrm{E}_{3D}$, which operates solely on the native coordinates of the 3D point cloud. In this paradigm, $\mathcal{F}_{2D}$ acts as the ``teacher'', while $\mathrm{E}_{3D}$ functions as the ``student''.

Specifically, the 3D encoder $\mathrm{E}_{3D}$ maps the input point cloud $P$ to per-point 3D feature embeddings $\mathcal{F}_{3D} = \{ \mathbf{f}_{3D,p} \mid \mathbf{p} \in \mathcal{P} \}$. To transfer knowledge from the teacher to the student, we minimize the cosine similarity loss $\mathcal{L}_{\text{distill}}$ between $\mathcal{F}_{3D}$ and $\mathcal{F}_{2D}$:
\begin{equation}
    \mathcal{L}_{\text{distill}} = 1 - \cos\big(\mathcal{F}_{3D}, \text{stop\_grad}(\mathcal{F}_{2D})\big),
\end{equation}
where the \texttt{stop\_grad} operation ensures that gradients are backpropagated only to the 3D encoder, updating only the student network's parameters.

Since the teacher features $\mathcal{F}_{2D}$ originate from a vision-language model aligned with CLIP, this distillation process ensures that the 3D features $\mathcal{F}_{3D}$ produced by the student network are also aligned with CLIP's text embedding space. This property enables the model to perform semantic segmentation at inference time using only the 3D point cloud and an arbitrary text prompt, without any reliance on 2D images.

\subsection{Inference with Hybrid 2D-3D Feature Fusion}
As highlighted in purple background in Figure~\ref{fig:OpenUrban3D_pipline}, at the inference stage we devise a hybrid feature fusion strategy to combine the complementary advantages of 2D and 3D features: 

\begin{itemize}
    \item \textbf{3D Features ($\mathcal{F}_{3D}$):} The distilled 3D features are imbued with rich geometric priors, demonstrating excellent performance in distinguishing between classes with significant structural differences (e.g., houses vs.\ ground, trees). They also tend to produce cleaner segmentation boundaries.

    \item \textbf{2D Features ($\mathcal{F}_{2D}$):} Originating from the vision-language model, the 2D features possess superior semantic recognition capabilities and are particularly adept at capturing small objects. However, due to point-cloud-to-pixel projection errors, their predictions are prone to noise at object boundaries.
\end{itemize}

We first evaluated a result-level fusion strategy, which involves separately calculating the similarity scores of $\mathcal{F}_{3D}$ and $\mathcal{F}_{2D}$ with the text embeddings, and then taking the maximum score as the final result. However, this method is highly dependent on both branches providing high-confidence predictions. In large-scale urban scenes, the prediction confidence of $\mathcal{F}_{3D}$ is generally low, causing this fusion strategy to sometimes degrade performance.

Therefore, we adopt a more direct feature-level fusion. We combine the 2D and 3D features for each point $\mathbf{p}$ using a weighted average:
\begin{equation}
    \label{eq3}
    \mathcal{F}_{\text{fusion}} = \alpha \cdot \mathcal{F}_{3D} + (1 - \alpha) \cdot \mathcal{F}_{2D},
\end{equation}
where $\alpha$ is a hyperparameter that balances the contribution of the two feature types. A key experimental finding is that the model's performance is optimal when $\alpha$ is set to a very small value. This suggests that the primary role of $\mathcal{F}_{3D}$ is to act as a geometric regularizer, injecting a small yet crucial amount of structural information into the semantically rich $\mathcal{F}_{2D}$ to correct its boundary noise, rather than dominating the final semantic decision.

Finally, the fused feature for each point, $\mathcal{F}_{\text{fusion}} = \{ \mathbf{f}_{\text{fusion},p} \mid \mathbf{p} \in \mathcal{P} \}$, is used for the final open-vocabulary semantic segmentation. By calculating its cosine similarity with the text embeddings $\{\mathbf{t}_n\}$ generated by CLIP for each class, we assign the semantic label to the point:
\begin{equation}
    \label{eq4}
    \text{Prediction} = \arg\max_{n} \cos(\mathcal{F}_{\text{fusion}}, \mathbf{t}_n).
\end{equation}

\subsection{LLM-Powered Text Queries}
The open-vocabulary nature of our model provides a natural interface for synergistic operation with Large Language Models (LLMs), significantly enhancing the system's interactive intelligence and flexibility. Users are no longer required to provide precise category names; instead, they can input complex, function-oriented, or intent-driven natural language commands. In this process, the LLM functions as a semantic parser, responsible for decomposing and mapping these high-level user queries into a concrete list of target categories for segmentation.

For example, given a complex command such as ``\texttt{Find all the places where people can walk and cars can drive}'', the LLM can intelligently parse this into a list of atomic categories, such as \texttt{[sidewall, road]}. This list is then passed to our model to perform precise open-vocabulary segmentation, thereby achieving a more intelligent and comprehensive understanding of the scene.

\begin{figure}
    \centering
    \includegraphics[width=1\linewidth]{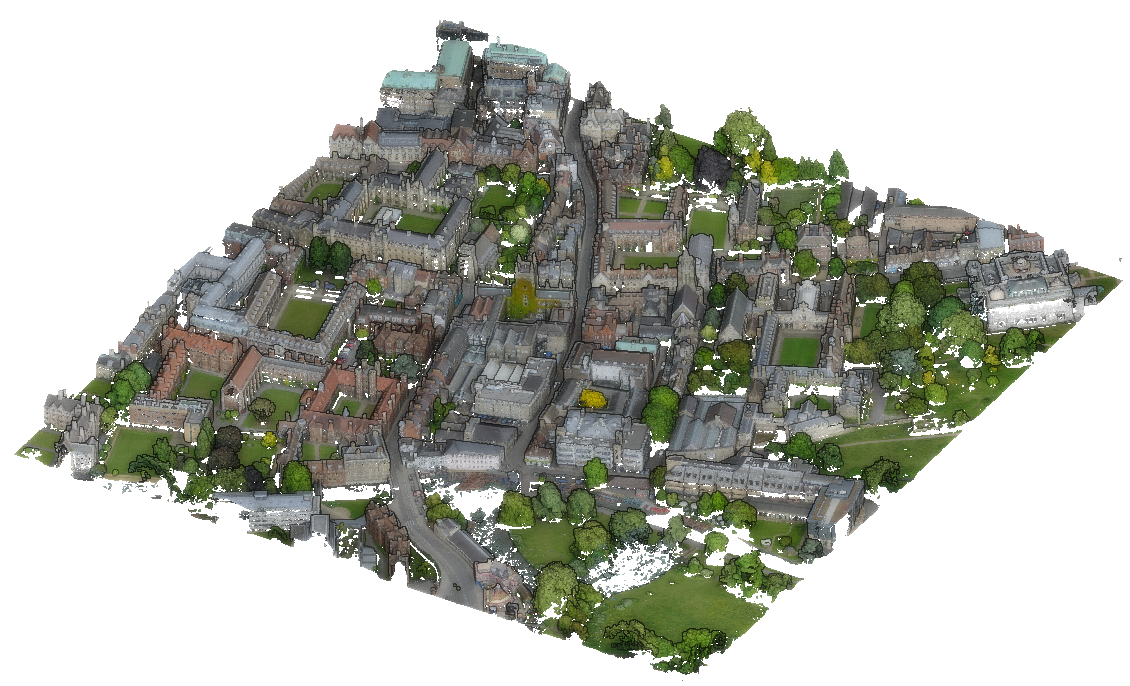}
    \caption{Example visualization from the SensatUrban dataset.}
    \label{fig:sensat_example}
\end{figure}

\begin{figure}
    \centering
    \includegraphics[width=0.9\linewidth]{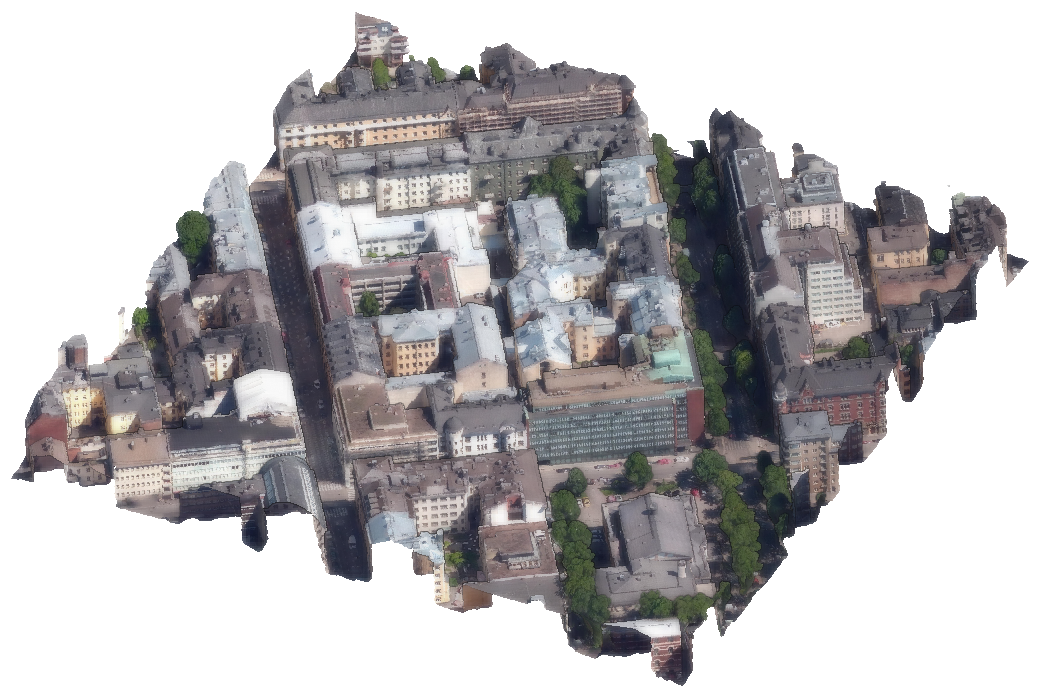}
    \caption{Example visualization from the SUM dataset.}
    \label{fig:sum_example}
\end{figure}

\section{Experiments and Analysis}
\subsection{Experimental Setup}

\subsubsection{Datasets}
We conduct our experiments on two widely used large-scale urban 3D semantic segmentation datasets: SensatUrban and SUM.

\textbf{SensatUrban} is an urban-scale photogrammetry point cloud dataset captured by Unmanned Aerial Vehicles (UAVs). The dataset is derived from aerial surveys of three cities in the UK (Birmingham, Cambridge, and York), covering an area of 7.6~km$^2$ and containing nearly three billion points with fine-grained semantic annotations. An example from the dataset is shown in Figure~\ref{fig:sensat_example}.

\textbf{The SUM  dataset} is constructed from a high-resolution 3D city mesh model of Helsinki, Finland. The original model was generated using oblique aerial photography. The dataset provides two point cloud versions sampled from the mesh model, with densities of 30~points/m$^2$ and 300~points/m$^2$. In our experiments, we use the version sampled at 30~points/m$^2$. An example is illustrated in Figure~\ref{fig:sum_example}.

\subsubsection{Experimental Setup and Metrics}
In the training phase of all our experiments, we follow an unsupervised learning paradigm and do not use any semantic labels from the datasets. The ground-truth labels are exclusively used during the performance evaluation phase to validate the effectiveness of our method.
To comprehensively evaluate the model's performance, we employ the following three standard metrics:

\textbf{Mean Intersection over Union (mIoU):}
\begin{equation}
    \text{mIoU} = \frac{1}{C} \sum_{i=1}^{C} \frac{\text{TP}_i}{\text{TP}_i + \text{FP}_i + \text{FN}_i}
\end{equation}
where $C$ is the total number of classes, and $\text{TP}_i$, $\text{FP}_i$, and $\text{FN}_i$ represent the number of true positive, false positive, and false negative predictions for class $i$, respectively.

\textbf{Mean Accuracy (mAcc):}
\begin{equation}
    \text{mAcc} = \frac{1}{C} \sum_{i=1}^{C} \frac{\text{TP}_i}{\text{TP}_i + \text{FN}_i}
\end{equation}

\textbf{Overall Accuracy (OA):}
\begin{equation}
    \text{OA} = \frac{\sum_{i=1}^{C} \text{TP}_i}{\sum_{i=1}^{C} (\text{TP}_i + \text{FN}_i)}
\end{equation}

\subsubsection{Implementation Details}

Our method is implemented based on the Python and PyTorch frameworks, and trained on an Ubuntu 18.04 server equipped with two NVIDIA A6000 GPUs. For the model architecture, we employ MinkUNet \cite{choy20194d} as the 3D backbone. We use the pre-trained and frozen Image Encoder from ODISE \cite{ODISE} as the 2D feature extractor and the pre-trained and frozen Text Encoder from CLIP \cite{clip} as the text feature extractor. During the 2D-3D distillation stage, we set the voxel size to 0.2 m and the batch size to 2 for all datasets, training for 60 epochs. We use the Adam optimizer for model optimization, with an initial learning rate set to 0.0001, which is gradually decreased as the number of iterations increases.

\subsection{Results}

This section presents the results of our OpenUrban3D model on the SensatUrban and SUM datasets. To comprehensively evaluate its performance, we compare it against two categories of baseline methods: (1) state-of-the-art open-vocabulary semantic segmentation methods, including Openscene \cite{peng2023openscene}, PLA \cite{ding2023pla}, and RegionPLC \cite{yang2024regionplc}; and (2) representative fully-supervised semantic segmentation methods.

\begin{table*}
    \caption{SensatUrban semantic segmentation results. Best results are marked in bold for both fully-supervised and open-vocabulary methods.}
    \begin{center}
        \resizebox{\textwidth}{!}{
            \begin{tabular}{@{}lllcccccccccccccc@{}}
                \hline
                Method                               & \rotatebox{45}{mIoU ($\% $)}
                                                     & \rotatebox{45}{mAcc($\% $)}
                                                     & \rotatebox{45}{OA ($\% $)}   & \rotatebox{45}{IoUs ($\% $)} &                      &                          &                      &                        &                         &                      &                          &                         &                     &                          &                      &                                                       \\ \cline{5-16}

                                                     &
                                                     &
                                                     &                              & \rotatebox{45}{ground}       & \rotatebox{45}{veg.} & \rotatebox{45}{building} & \rotatebox{45}{wall} & \rotatebox{45}{bridge} & \rotatebox{45}{parking} & \rotatebox{45}{rail} & \rotatebox{45}{traffic.} & \rotatebox{45}{street.} & \rotatebox{45}{car} & \rotatebox{45}{footpath} & \rotatebox{45}{bike} & \rotatebox{45}{water}                                 \\
                \rowcolor{gray!25} 
                \multicolumn{17}{l}{\textbf{Fully-supervised methods}}                                                                                                                                                                                                                                                                                                                                                                                     \\ 
                \hline
                PointNet \cite{pointnet}             & 23.7
                                                     & -
                                                     & 80.7                         & 67.9                         & 89.5                 & 80.0                     & 0.0                  & 0.0                    & 3.9                     & 0.0                  & 31.5                     & 0.0                     & 35.1                & 0.0                      & 0.0                  & 0.0                                                   \\

                PointNet++ \cite{pointnet++}         & 32.9
                                                     & -
                                                     & 84.3                         & 72.4                         & 94.2                 & 84.7                     & 2.7                  & 2.0                    & 25.7                    & 0.0                  & 31.5                     & 11.4                    & 38.8                & 7.1                      & 0.0                  & 56.9                                                  \\

                SPGraph \cite{spg}                   & 37.2
                                                     & -
                                                     & 76.9                         & 69.9                         & 94.5                 & 88.8                     & 32.8                 & 12.5                   & 15.7                    & 15.4                 & 30.6                     & 22.9                    & 56.4                & 0.5                      & 0.0                  & 44.2                                                  \\

                SparseConv \cite{sparseconv}         & 42.6
                                                     & -
                                                     & 85.2                         & 74.1                         & 97.9                 & 94.2                     & 63.3                 & 7.5                    & 24.2                    & 0.0                  & 30.1                     & 34.0                    & 74.4                & 0.0                      & 0.0                  & 54.8                                                  \\
                MinkUnet  ~\cite{choy20194d}         & 46.2
                                                     & 53.3
                                                     & 90.8                         & 81.5                         & 97.7                 & 94.8                     & 59.6                 & 0.1                    & 30.7                    & 0.0                  & 55.0                     & 33.1                    & 77.0                & 22.9                     & 0.0                  & 49.1                                                  \\
                RandLA-Net~\cite{randla}             & 52.7                         & -                            & 89.8                 & 80.1                     & 98.1                 & 91.6                   & 48.9                    & 40.8                 & 51.6                     & 0.0                     & 56.7                & 33.3                     & 80.1                 & 32.6                  & 0.0           & 71.3
                \\
                KPConv ~\cite{kpconv}                & 57.6                         & -                            & 93.2                 & 87.1                     & 98.3                 & 95.3                   & 74.4                    & 28.7                 & 41.4                     & 0.0                     & 56.0                & 54.4                     & 85.7                 & 40.4                  & 0.0           & \textbf{86.3}
                \\
                Point Transformer~\cite{ptv1}        & 61.8                         & -                            & 92.4                 & 84.8                     & 98.6                 & 95.2                   & 61.9                    & \textbf{64.6}        & 55.4                     & 22.3                    & 57.7                & 43.8                     & 81.1                 & 42                    & 18.9          & 76.6
                \\
                LACV-Net~\cite{lacvnet}              & 61.3                         & -                            & 93.2                 & 85.5                     & 98.4                 & \textbf{95.6}          & 61.9                    & 58.6                 & \textbf{64.0}            & 28.5                    & 62.8                & 45.4                     & 81.9                 & 42.4                  & 4.8           & 67.7
                \\
                PointNAT~\cite{zeng2024pointnat}     & 65.2                         & -                            & 93.8                 & 85.6                     & \textbf{98.9}        & 96.4                   & \textbf{77.7}           & 57.6                 & 54.5                     & 52.6                    & \textbf{64.3}       & \textbf{57.5}            & \textbf{86.4}        & 38.5                  & 0.0           & 78.1
                \\
                EyeNet++~\cite{yoo2025eyenet++}      & \textbf{68.2}                & -                            & \textbf{93.9}        & \textbf{87.6}            & 98.7                 & 96.2                   & 69.1                    & 62.0                 & 58.7                     & \textbf{53.7}           & 62.1                & 53.5                     & 84.0                 & \textbf{47.1}         & \textbf{30.0} & 84.3
                \\
                \rowcolor{gray!25}
                \multicolumn{17}{l}{\textbf{Open-Vocabulary methods}}                                                                                                                                                                                                                                                                                                                                                                                      \\
                \hline
                OpenScene  ~\cite{peng2023openscene} & 12.6
                                                     & 21.4
                                                     & 51.6                         & 41.8                         & 55.6                 & 39.1                     & 0.0                  & 0.0                    & 0.0                     & 0.0                  & 18.3                     & 0.0                     & 1.9                 & 0.8                      & 0.0                  & 6.0                                                   \\
                PLA ~\cite{ding2023pla}              & 1.7
                                                     & 6.1
                                                     & 17.6                         & 21.6                         & 0.0                  & 0.0                      & 0.0                  & 0.0                    & \textbf{0.1}            & 0.0                  & 0.0                      & 0.0                     & 0.0                 & 0.0                      & 0.0                  & 0.0                                                   \\
                RegionPLC  ~\cite{yang2024regionplc} & 2.5
                                                     & 8.6
                                                     & 25.2                         & 28.1                         & 0.0                  & 2.0                      & \textbf{1.2}         & 0.0                    & 0.0                     & 0.0                  & 0.2                      & 0.0                     & 1.3                 & 0.1                      & 0.0                  & 0.0                                                   \\

                OpenUrban3D                          & \textbf{39.6}                & \textbf{46.2}                & \textbf{84.7}        & \textbf{70.8}            & \textbf{90.8}        & \textbf{85.8}          & 0.2                     & \textbf{85.2}        & 0.0                      & 0.0                     & \textbf{46.4}       & 0.0                      & \textbf{72.6}        & \textbf{8.7}          & 0.0           & \textbf{53.7}
                \\
                \hline
            \end{tabular}
        }
    \end{center}
    \label{tab:one}
\end{table*}

\begin{figure*}[t]
    \centering
    \includegraphics[height=16cm, keepaspectratio]{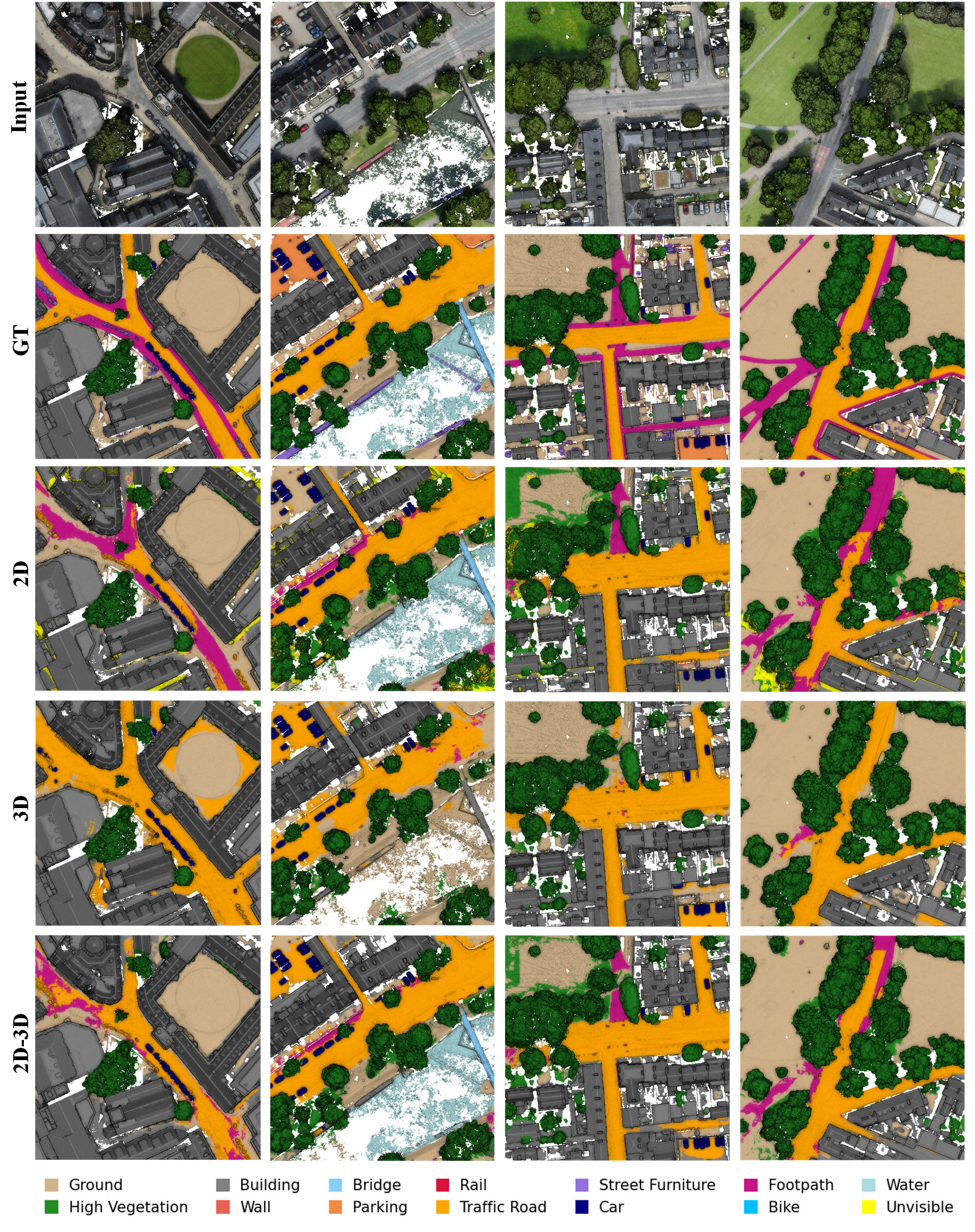}
    \caption{Qualitative visualization results from OpenUrban3D on the SensatUrban dataset.}
    \label{fig:sensaturban_qualit}
\end{figure*}

\subsubsection{\textbf{SensatUrban}}

As presented in Table~\ref{tab:one} and Figure~\ref{fig:sensaturban_qualit}, we conducted both quantitative and qualitative evaluations of the OpenUrban3D model on the SensatUrban dataset. Quantitatively, OpenUrban3D achieved an mIoU of 39.6\% and an OA of 84.7\%, far surpassing existing 3D open-vocabulary segmentation models (with the best results being 12.6\% mIoU and 51.6\% OA for OpenScene). PLA and RegionPLC almost completely failed to produce reasonable results. These findings demonstrate that many prior methods, particularly those primarily designed for indoor scenes or relying on ideal image-text pairs, struggle to generalize directly to large-scale outdoor environments, especially when confronted with real-world, multi-view imagery lacking precise alignment.

Compared to fully supervised methods, the current state-of-the-art supervised approach, EyeNet++, achieved an mIoU of 68.2\%. The fully supervised results from MinkUNet, which serves as the 3D backbone in OpenUrban3D, reached an mIoU of 46.2\%, only 6.6\% higher than our model. This narrow gap highlights the strong capability of OpenUrban3D to learn discriminative and semantically rich 3D representations in an open-vocabulary setting, despite the absence of dense per-point supervision.

On a per-class basis, the model performs exceptionally well on categories with clear and unambiguous semantics, such as buildings and vegetation. However, its performance is weaker on classes like ``wall'' and various street facilities. Our analysis suggests this is primarily due to the high intra-class diversity within the SensatUrban dataset. For instance, the ``wall'' category may include fences, barriers, and retaining walls, while street facilities can encompass a wide range of objects like street lamps, traffic signs, and trash cans. A single text description struggles to accurately capture such a broad semantic range. For other categories like ``rail'' and ``bike'', the primary cause of failure is the scarcity of corresponding points in the dataset. This sparsity makes them difficult to recognize effectively in the projected 2D rendered images, thereby limiting the model's learning capability.

\begin{table*}
    \caption{SUM semantic segmentation results.  Best results are marked in bold for both fully-supervised and open-vocabulary methods.}
    \begin{center}
        \begin{tabular}{@{}lllccccccc}
            \hline
            Method                              & \rotatebox{45}{mIoU ($\% $)}
                                                & \rotatebox{45}{mAcc ($\% $)}
                                                & \rotatebox{45}{OA ($\% $)}   & \rotatebox{45}{IoUs ($\% $)} &                &               &               &               &                                               \\ \cline{5-10}

                                                &                              &                              &                & Terrain       & High
            Vegetation
                                                & Building                     & Water                        & Vehicle        & Boat                                                                                          \\
            \rowcolor{gray!25} 
            \multicolumn{10}{l}{\textbf{Fully-supervised methods}}                                                                                                                                                             \\ 
            \hline
            PointNet \cite{pointnet}            & 23.7
                                                & -
                                                & 80.7                         & 67.9                         & 89.5           & 80.0          & 0.0           & 0.0           & 3.9                                           \\

            PointNet++ \cite{pointnet++}        & 32.9
                                                & -
                                                & 84.3                         & 72.4                         & 94.2           & 84.7          & 2.7           & 2.0           & 25.7                                          \\

            SPGraph \cite{spg}                  & 37.2
                                                & -
                                                & 76.9                         & 69.9                         & 94.5           & 88.8          & 32.8          & 12.5          & 15.7                                          \\

            SparseConv \cite{sparseconv}        & 42.6
                                                & -
                                                & 85.2                         & 74.1                         & \textbf{97.9}  & \textbf{94.2} & 63.3          & 7.5           & 24.2                                          \\
            MinkUnet~\cite{choy20194d}          & 70.0                         & 74.5                         & 93.3           & 84.4          & 92.0          & 92.9          & 76.8          & \textbf{61.6} & 12.6          \\
            KPConv~\cite{kpconv}                & 68.8
                                                & -                            & 94.2
                                                & 86.5                         & 88.4                         & 92.7           & 77.7          & 54.3          & 13.3
            \\
            RandLaNet~\cite{randla}             & 38.6
                                                & -                            & 74.9
                                                & 38.9                         & 59.6                         & 81.5           & 27.7          & 22.0          & 2.1
            \\
            PTV2~\cite{ptv2}                    & 65.4
                                                & -                            & 93.1
                                                & 82.8                         & 89.6                         & 93.6           & 87.4          & 19.8          & 18.8
            \\
            PTV3~\cite{ptv3}                    & \textbf{74.0}
                                                & -                            & \textbf{94.7}
                                                & \textbf{87.3}                & 92.3                         & \textbf{94.2}  & \textbf{88.5} & 49.4          & \textbf{32.4}
            \\
            \rowcolor{gray!25} 
            \multicolumn{10}{l}{\textbf{Open-Vocabulary methods}}                                                                                                                                                              \\
            \hline
            OpenScene ~\cite{peng2023openscene} & 41.1
                                                & 46.5
                                                & 74.6                         & 35.6                         & 54.1           & 73.7          & 53.9          & 0.1           & 28.8                                          \\
            PLA ~\cite{ding2023pla}             & 4.3
                                                & 16.7
                                                & 26.7                         & 24.6                         & 0.0            & 0.2           & 0.0           & 0.9           & 0.0                                           \\
            RegionPLC~\cite{yang2024regionplc}  & 8.1
                                                & 16.6
                                                & 26.4                         & 16.2                         & 7.5            & 19.3          & 3.1           & 0.1           & 2.6                                           \\

            OpenUrban3D                         & \textbf{75.4}                & \textbf{86.8}                & \textbf{90.5} & \textbf{69.3} & \textbf{83.6} & \textbf{90.2} & \textbf{92.7} & \textbf{48.3} & \textbf{68.1}
            \\
            \hline
        \end{tabular}
    \end{center}
    \label{tab:two}
\end{table*}

\begin{figure*}[t]
    \centering
    \includegraphics[height=16cm, keepaspectratio]{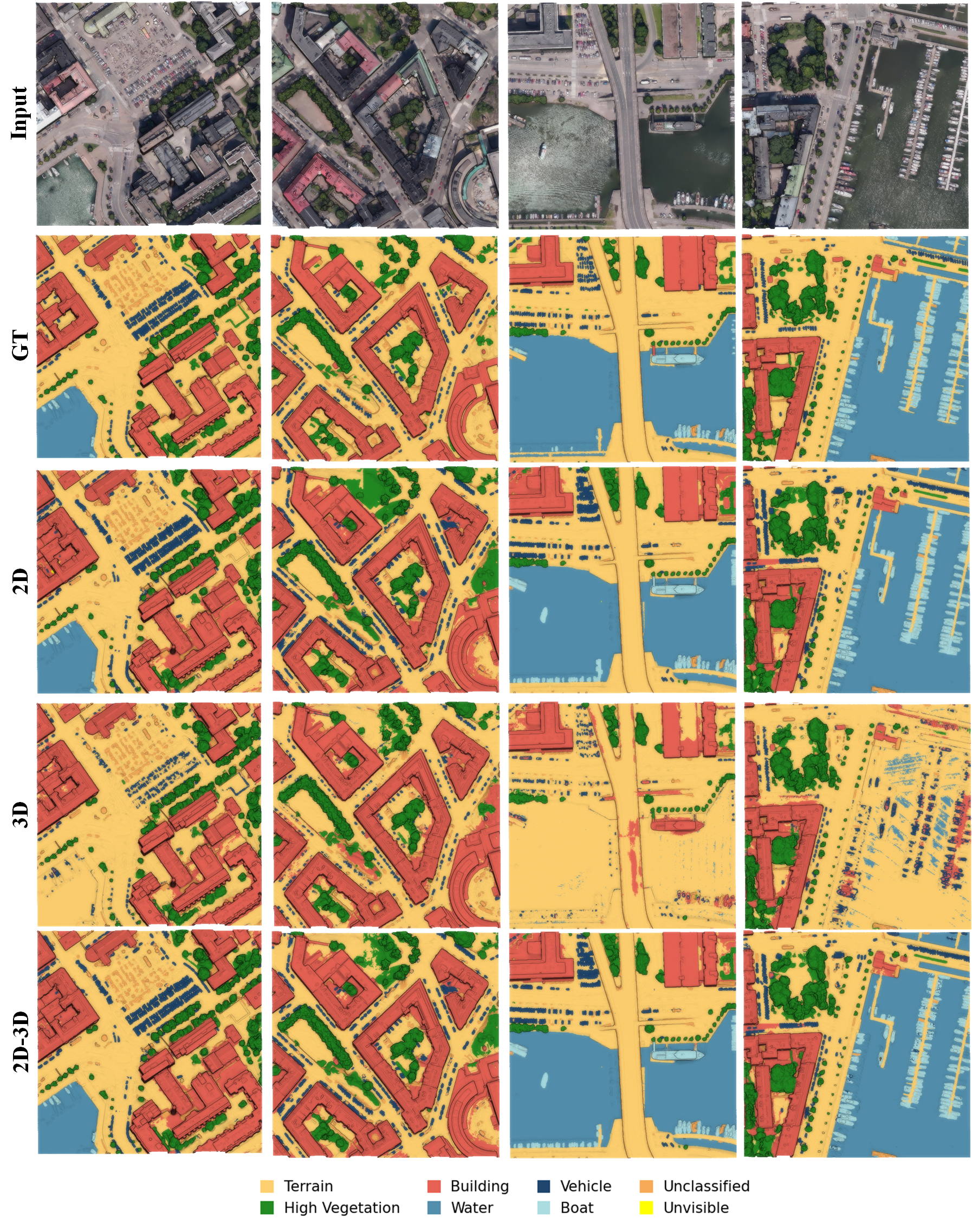}
    \caption{Qualitative visualization results from OpenUrban3D on the SUM dataset.}
    \label{fig:SUM_qualit}
\end{figure*}

\subsubsection{\textbf{SUM}}

As shown in Table~\ref{tab:two} and Figure~\ref{fig:SUM_qualit}, we evaluate OpenUrban3D on the SUM dataset through both quantitative and qualitative analyses. Our method achieves 75.4\% mIoU and 90.5\% OA, significantly outperforming existing open-vocabulary models, the best of which, OpenScene, reaches only 41.1\% mIoU. Similar to the results on SensatUrban, PLA and RegionPLC fail to produce meaningful outputs.

A particularly notable finding is that OpenUrban3D not only surpasses all open-vocabulary baselines, but also exceeds the performance of state-of-the-art fully supervised methods. For example, PTV3~\cite{ptv3}, the current top-performing supervised approach, achieves 74.0\% mIoU, which is lower than our result. This remarkable outcome suggests that, for large-scale urban datasets with well-defined categories such as SUM dataset, the generalized visual knowledge distilled from large-scale, pre-trained 2D foundation models can be more effective than direct supervised training on limited 3D data.

\begin{figure}[t]
    \centering
    \includegraphics[width=\columnwidth]{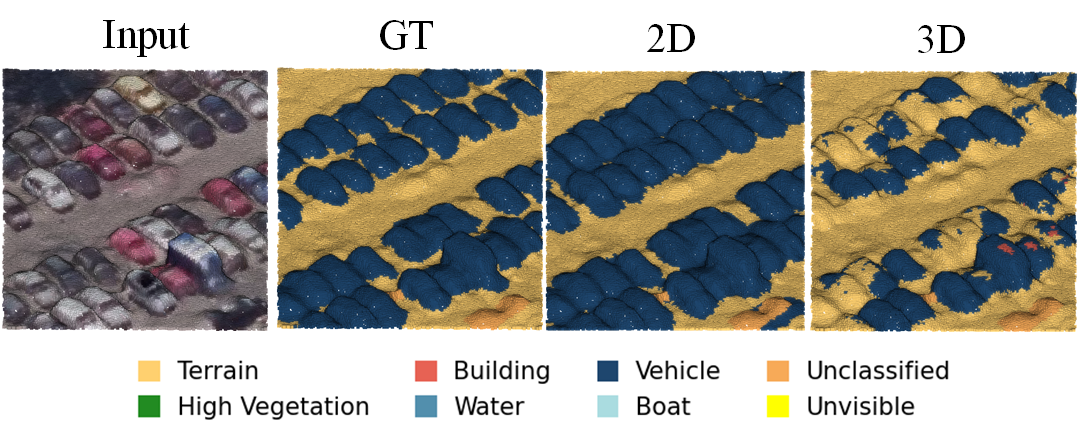}
    \caption{Limitations of 3D-only features in distinguishing geometrically fused objects, exemplified by terrain and vehicles.}
    \label{fig:sum_3d_limitations}
\end{figure}

A deeper analysis of the per-class results shows that the performance of the pure 3D features declines significantly on the ``Terrain'' and ``Vehicle'' classes. As illustrated in Figure~\ref{fig:sum_3d_limitations}, we attribute this to the fact that these categories lack unique, distinguishable geometric shapes, for instance, the transition between vehicles and the ground is relatively smooth, making vehicles appear as only slight protrusions on the ground in terms of geometric features. This again highlights the critical importance of 2D semantic features for recognizing such objects. Although our final fused model performs excellently on these classes, this analysis reveals the inherent limitations of purely geometric models in semantic recognition and, in turn, underscores the necessity and sophistication of our 2D-3D fusion strategy.

\subsection{Analysis on Different Feature Combinations}
To deeply analyze the effectiveness of the different feature components in our model, we further report the results under the following four variant configurations:

\begin{table*}[]
    \centering
    \caption{Segmentation results with different feature combinations.}
    \label{tab:featureablation}
    \begin{tabular}{lcccccc}
        \hline
                                 & \multicolumn{3}{c}{SensatUrban} & \multicolumn{3}{c}{SUM}                                             \\ \cline{2-7}
        Feature option           & mIoU (\%)                       & mAcc (\%)               & OA (\%) & mIoU (\%) & mAcc (\%) & OA (\%) \\ \hline
        OpenUrban3D (2D Feature) & 38.4                            & 46.7                    & 82.9    & 77.2      & 85.7      & 95.7    \\
        OpenUrban3D (3D Feature) & 28.2                            & 33.5                    & 84.7    & 45.4      & 53.4      & 84.7    \\
        OpenUrban3D (2D-3D vis)  & 40.1                            & 46.8                    & 83.9    & 78.5      & 89.5      & 91.2    \\
        OpenUrban3D (2D-3D)      & 39.6                            & 46.2                    & 84.7    & 75.4      & 86.8      & 90.5    \\ \hline
    \end{tabular}
\end{table*}

\begin{table*}
    \caption{Ablation studies on the SBFF module and Text Prompt.}
    \begin{center}
        \resizebox{\textwidth}{!}{
            \begin{tabular}{@{}lllcccccccccccccc@{}}
                \hline
                Method                                  & \rotatebox{45}{mIoU ($\% $)}
                                                        & \rotatebox{45}{mAcc($\% $)}
                                                        & \rotatebox{45}{OA ($\% $)}   & \rotatebox{45}{IoUs ($\% $)} &                      &                          &                      &                        &                         &                      &                          &                         &                     &                          &                      &                                    \\ \cline{5-17}

                                                        &
                                                        &
                                                        &                              & \rotatebox{45}{ground}       & \rotatebox{45}{veg.} & \rotatebox{45}{building} & \rotatebox{45}{wall} & \rotatebox{45}{bridge} & \rotatebox{45}{parking} & \rotatebox{45}{rail} & \rotatebox{45}{traffic.} & \rotatebox{45}{street.} & \rotatebox{45}{car} & \rotatebox{45}{footpath} & \rotatebox{45}{bike} & \rotatebox{45}{water}              \\

                \rowcolor{gray!25} 
                \multicolumn{17}{l}{\textbf{Ablation study on the SBFF module}}                                                                                                                                                                                                                                                                                                                                                            \\
                \hline
                OpenUrban3D (3D Features) w/o SBFF      & 27.9                         & 32.8                         & 85.0                 & 69.7                     & 94.5                 & 88.5                   & 0.0                     & 0.0                  & 0.0                      & 0.0                     & 41.0                & 0.0                      & 69.2                 & 0.0                   & 0.0 & 0.0  \\
                OpenUrban3D (3D Features) w/ SBFF       & 28.2
                                                        & 33.5
                                                        & 84.7                         & 70.0                         & 94.7                 & 87.6                     & 0.0                  & 0.0                    & 0.0                     & 0.0                  & 41.2                     & 0.0                     & 72.0                & 1.4                      & 0.0                  & 0.0                                \\
                \rowcolor{gray!25} 
                \multicolumn{17}{l}{\textbf{Alation study on text prompt}}                                                                                                                                                                                                                                                                                                                                                                 \\
                \hline
                OpenUrban3D (2D-3D) w/ Original Prompt  & 30.7                         & 39.5
                                                        & 72.5                         & 70.4                         & 91.3                 & 73.1                     & 2.9                  & 15.6                   & 0.0                     & 0.1                  & 12.0                     & 0.0                     & 69.6                & 8.8                      & 0.0                  & 55.8
                \\
                OpenUrban3D (2D-3D) w/ Processed Prompt & 39.6                         & 46.2                         & 84.7                 & 70.8                     & 90.8                 & 85.8                   & 0.2                     & 85.2                 & 0.0                      & 0.0                     & 46.4                & 0.0                      & 72.6                 & 8.7                   & 0.0 & 53.7
                \\
                \hline
            \end{tabular}
        }
    \end{center}
    \label{tab:ab_sbff_text}
\end{table*}
\begin{itemize}
    \item \textbf{OpenUrban3D (2D Features)}: Evaluation is performed using only the features ($F_\text{2D}$) extracted from multi-view 2D images. Since not all 3D points are visible in 2D images, the evaluation metrics are computed only on the 2D-visible points.

    \item \textbf{OpenUrban3D (3D Features)}: Evaluation is based solely on the features ($F_\text{3D}$) derived from the 3D model distilled from 2D features, covering all points.

    \item \textbf{OpenUrban3D (2D-3D vis)}: Evaluation uses the fused 2D-3D features but is limited to the 2D-visible points, allowing a fair comparison with \textbf{OpenUrban3D (2D Features)} and demonstrating the contribution of 3D features.

    \item \textbf{OpenUrban3D (2D-3D)}: This is our complete proposed model. It utilizes the fused 2D-3D features for evaluation. For points not visible in the 2D views, the model relies solely on the 3D features ($F_\text{3D}$). The evaluation is conducted on all points.
\end{itemize}

\begin{figure}[t]
    \centering
    \includegraphics[width=\columnwidth]{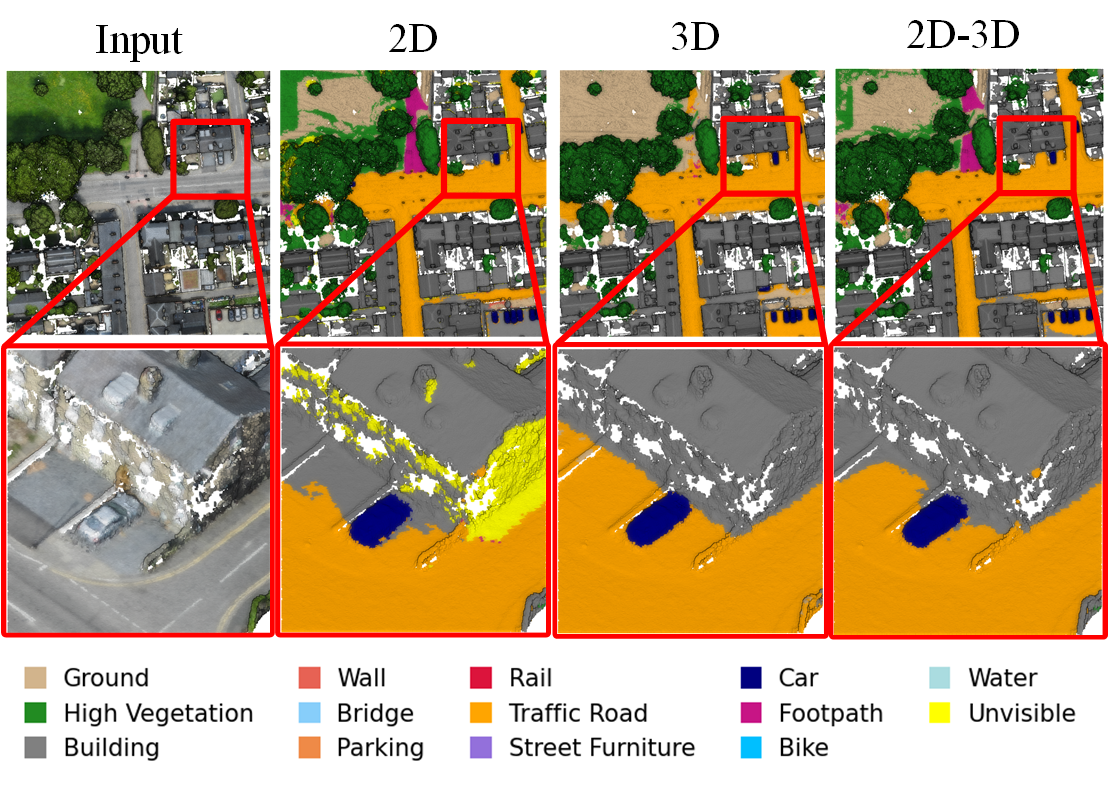}
    \caption{Comparison of different feature combinations for semantic segmentation in the SensatUrban dataset.}
    \label{fig:geometric_edges_sensaturban}
\end{figure}

\begin{figure}[t]
    \centering
    \includegraphics[width=\columnwidth]{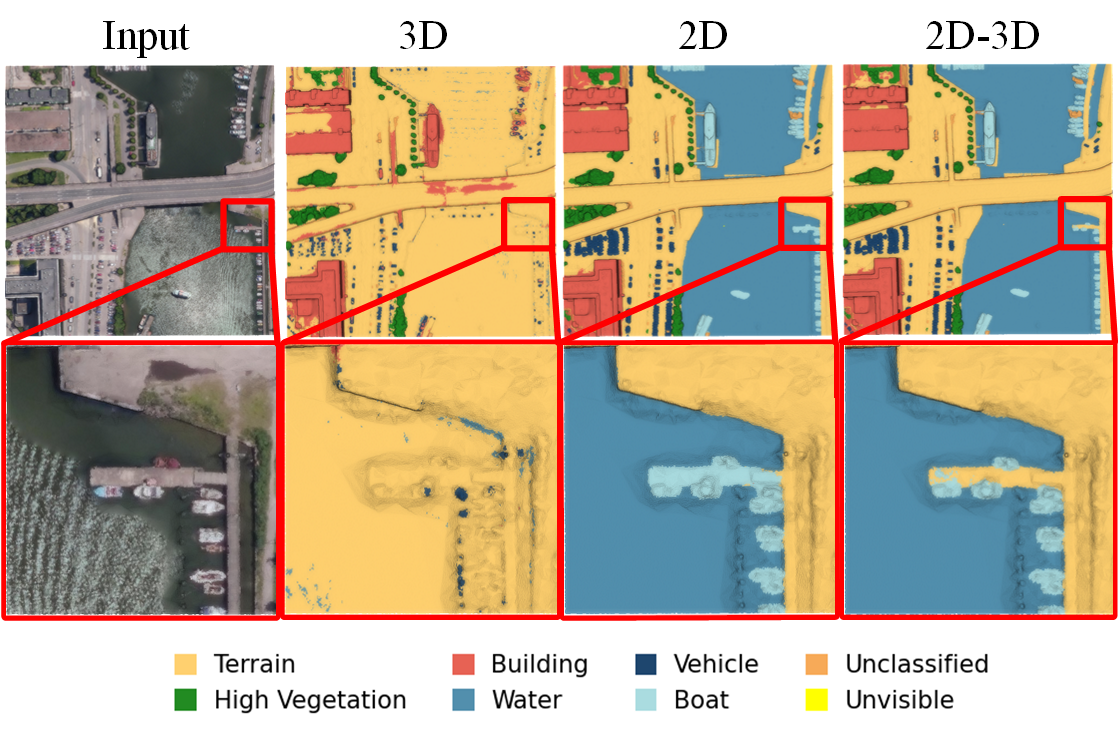}
    \caption{Comparison of different feature combinations for semantic segmentation in the SUM dataset.}
    \label{fig:geometric_edges_sum}
\end{figure}

By comparing the performance of different feature combinations on both the SensatUrban and SUM datasets, we gain deeper insights into the model's internal working mechanisms. As shown in Table \ref{tab:featureablation}, in both cases, the 2D features provide a strong semantic foundation (38.4\% mIoU on SensatUrban and 77.2\% on SUM when used alone), forming the cornerstone of the model's semantic understanding, while the 3D features serve as a crucial geometric supplement. On SensatUrban, combining 2D and 3D features improves mIoU from 38.4\% to 39.6\%, with the purely 3D model producing sharper and more precise segmentation along object boundaries, such as more accurately delineating the boundaries between buildings and the ground (Figure~\ref{fig:geometric_edges_sensaturban}). A notable observation is that the final model’s performance on the entire point cloud (39.6\%) is nearly identical to that on only the 2D-visible points (40.1\%), indicating that the 3D branch robustly extends coverage from visible-only points to the entire 3D scene without significant degradation. On SUM, pure 3D features outperform 2D features in categories with distinct geometric profiles, such as ``High Vegetation'' and ``Building'', highlighting their advantage in capturing precise geometry and refining boundaries. The fused features achieve the best overall performance (mIoU 78.5\%), as exemplified by the ``Boat'' class, where IoU rises from 44\% with 2D features alone and 0\% with 3D features (which lack semantic recognition) to 83.6\% after fusion (Figure~\ref{fig:geometric_edges_sum}). These results demonstrate that the proposed fusion strategy effectively integrates the semantic recognition capability of 2D features with the geometric refinement provided by 3D features across diverse large-scale urban datasets.

\begin{table}[htbp]
    \centering
    \caption{Ablation study on $A$ and $R$ under different $K$ values for Multi-view Multi-granularity Projection,where $S\_R$ indicates the proportion of scene points observed by the camera.}
    \resizebox{\linewidth}{!}{%
        \begin{tabular}{cc}
            \begin{minipage}{0.46\linewidth}
                \centering
                \setlength{\tabcolsep}{5.5pt}  
                \textbf{$K=2$} \\[0.6ex]
                \begin{tabular}{c c c c}
                    \toprule
                    \textbf{$A$} & \textbf{$R$} & \textbf{$S\_R$ (\%)} & \textbf{mIoU} \\
                    \midrule
                    $90^\circ$   & 0.5          & 94.2                 & 74.4          \\
                                 & 1            & 85.3                 & 73.8          \\
                                 & 2            & 80.0                 & 73.3          \\
                    \cmidrule(lr){1-4}
                    $120^\circ$  & 0.5          & 92.2                 & 74.1          \\
                                 & 1            & 83.5                 & 73.5          \\
                                 & 2            & 78.8                 & 73.2          \\
                    \cmidrule(lr){1-4}
                    $180^\circ$  & 0.5          & 88.1                 & 73.6          \\
                                 & 1            & 80.1                 & 72.5          \\
                                 & 2            & 76.3                 & 72.3          \\
                    \bottomrule
                \end{tabular}
            \end{minipage}
             &
            \begin{minipage}{0.46\linewidth}
                \centering
                \setlength{\tabcolsep}{5.5pt}  
                \textbf{$K=3$} \\[0.6ex]
                \begin{tabular}{c c c c}
                    \toprule
                    \textbf{$A$} & \textbf{$R$} & \textbf{$S\_R$ (\%)} & \textbf{mIoU} \\
                    \midrule
                    $90^\circ$   & 0.5          & 95.4                 & 76.6          \\
                                 & 1            & 88.9                 & 76.0          \\
                                 & 2            & 84.1                 & 75.3          \\
                    \cmidrule(lr){1-4}
                    $120^\circ$  & 0.5          & 94.1                 & 76.1          \\
                                 & 1            & 87.5                 & 75.9          \\
                                 & 2            & 83.2                 & 74.9          \\
                    \cmidrule(lr){1-4}
                    $180^\circ$  & 0.5          & 91.0                 & 75.6          \\
                                 & 1            & 84.6                 & 75.4          \\
                                 & 2            & 80.8                 & 74.3          \\
                    \bottomrule
                \end{tabular}
            \end{minipage}
            \\
            \addlinespace[2ex]
            \begin{minipage}{0.46\linewidth}
                \centering
                \setlength{\tabcolsep}{5.5pt}  
                \textbf{$K=4$} \\[0.8ex]
                \begin{tabular}{c c c c}
                    \toprule
                    \textbf{$A$} & \textbf{$R$} & \textbf{$S\_R$ (\%)} & \textbf{mIoU} \\
                    \midrule
                    $90^\circ$   & 0.5          & \textbf{95.7}        & \textbf{77.2} \\
                                 & 1            & 90.1                 & 76.3          \\
                                 & 2            & 86.1                 & 75.6          \\
                    \cmidrule(lr){1-4}
                    $120^\circ$  & 0.5          & 94.7                 & 76.9          \\
                                 & 1            & 89.0                 & 76.3          \\
                                 & 2            & 85.3                 & 75.6          \\
                    \cmidrule(lr){1-4}
                    $180^\circ$  & 0.5          & 92.3                 & 76.7          \\
                                 & 1            & 86.6                 & 75.8          \\
                                 & 2            & 83.1                 & 75.1          \\
                    \bottomrule
                \end{tabular}
            \end{minipage}
             &
            \begin{minipage}{0.46\linewidth}
                \centering
                \setlength{\tabcolsep}{5.5pt}  
                \textbf{$K=5$} \\[0.8ex]
                \begin{tabular}{c c c c}
                    \toprule
                    \textbf{$A$} & \textbf{$R$} & \textbf{$S\_R$ (\%)} & \textbf{mIoU} \\
                    \midrule
                    $90^\circ$   & 0.5          & 95.6                 & 76.8          \\
                                 & 1            & 90.6                 & 75.7          \\
                                 & 2            & 87.0                 & 74.6          \\
                    \cmidrule(lr){1-4}
                    $120^\circ$  & 0.5          & 94.7                 & 76.3          \\
                                 & 1            & 89.7                 & 74.9          \\
                                 & 2            & 86.3                 & 73.7          \\
                    \cmidrule(lr){1-4}
                    $180^\circ$  & 0.5          & 92.9                 & 75.5          \\
                                 & 1            & 87.7                 & 74.7          \\
                                 & 2            & 84.5                 & 73.8          \\
                    \bottomrule
                \end{tabular}
            \end{minipage}
        \end{tabular}
    }
    \label{tab:ab_study_MMP_sub}
\end{table}

\subsection{Ablation Studies}

\subsubsection{Multi-view Multi-granularity Projection Module}
We conducted an ablation study on the Multi-view Multi-granularity Projection module to investigate the impact of its key hyperparameters (as shown in Table~\ref{tab:ab_study_MMP_sub}). As defined in Section~\ref{sec:MMGP}, we evaluated three parameters: view granularity $K$, view sampling density $A$, and camera trajectory radius $R$. Specifically, $K$ determines the coverage area of the views and the number of anchors; a smaller $K$ results in fewer but more expansive views with sparser anchors. $A$ controls the density of viewpoint sampling, where a smaller value leads to richer sampled views. $R$ defines the camera’s shooting radius, where a smaller value produces a more oblique viewing angle, potentially causing more scene occlusion. Our experiments show that performance is highly sensitive to these parameters, with an optimal $K$ at 4. A $K$ that is too large (e.g., $K=5$) can degrade performance due to distortions and artifacts from point cloud rendering, while an appropriate $K$ balances fine-grained detail capture and artifact avoidance. We also find that denser view sampling (smaller $A$) and camera settings that reduce occlusion (smaller $R$) yield richer multi-view imagery and a higher $S\_R$, meaning that more points in the scene are captured by the cameras, improving segmentation accuracy. However, these optimal settings often incur higher computational costs, so a trade-off between accuracy and efficiency should be considered for practical deployment.

\subsubsection{Sample-Balanced Feature Fusion}

As shown in Table~\ref{tab:ab_sbff_text}, we conducted an ablation study on the Sample-Balanced Feature Fusion (SBFF) module to validate its effectiveness in promoting the learning of under-represented classes. The results clearly indicate that after incorporating the SBFF module, the model's mIoU and mAcc increased from 27.9\% and 32.8\% to 28.2\% and 33.5\%, respectively. This performance gain primarily stems from more effective learning of classes that are either sparse or small-sized. The most significant example is observed in the ``Footpath'' class: without the SBFF module, the model tended to misclassify footpaths as the visually similar ``Ground'' class due to sample imbalance, resulting in an IoU of 0\%. After applying SBFF, the model successfully discriminated between the two, achieving a breakthrough for the ``Footpath'' class with its IoU increasing from 0\% to 1.4\%. Furthermore, the module also enhanced the segmentation accuracy for small-sized objects like the ``Car'' class, with its IoU improving from 69.2\% to 72.0\%. These results provide compelling evidence that our sample-balancing strategy can effectively enhance the model's attention to and discriminative capability for classes that are either sparse in point count or easily confused, thereby improving the balance and robustness of the overall segmentation performance.

\begin{table}
    \caption{Ablation study on the impact of 2D-3D feature fusion strategies.}
    \centering
    \begin{tabular}{lccc}\toprule
                               & mIoU ($\% $) & mAcc ($\% $) & OA ($\% $) \\\midrule
        \rowcolor{gray!25}
        \multicolumn{4}{l}{SensatUrban}                                   \\
        \hline
        OpenUrban3D (Ensemble) & 38.9         & 46.6         & 83.5       \\
        OpenUrban3D (Fusion)   & 40.1
                               & 46.8
                               & 83.9                                     \\
        \rowcolor{gray!25}
        \multicolumn{4}{l}{SUM}                                           \\
        \hline
        OpenUrban3D (Ensemble)
                               & 72.6         & 83.9         & 89.8       \\
        OpenUrban3D (Fusion)   & 78.5         & 89.5         & 91.2       \\ \bottomrule
    \end{tabular}
    \label{tab:ab_fusion}
\end{table}

\subsubsection{Impact of Different 2D-3D Feature Fusion Strategies}
As shown in Table~\ref{tab:ab_fusion}, we conducted an ablation study on different strategies for fusing 2D and 3D features. \textbf{Ensemble} refers to separately computing the similarity scores of $\mathcal{F}_{2D}$ and $\mathcal{F}_{3D}$ with the text embeddings and then taking the maximum score as the final prediction, while \textbf{Fusion} refers to a weighted average of $\mathcal{F}_{2D}$ and $\mathcal{F}_{3D}$ followed by computing the similarity with the text features. Experimental results show that the Fusion strategy achieves superior performance on both datasets. This is because the Ensemble approach relies heavily on both $\mathcal{F}_{2D}$ and $\mathcal{F}_{3D}$ branches providing high-confidence predictions, whereas in large-scale scenes, the prediction confidence of F3D is generally low due to the larger voxel size. In contrast, the Fusion strategy can smoothly integrate the geometric information from $\mathcal{F}_{3D}$ into the final fused feature F-fusion. By assigning a relatively small weight to $\mathcal{F}_{3D}$, the method effectively extracts only the geometric priors from $\mathcal{F}_{3D}$, thereby improving overall performance.
\begin{figure}
    \centering
    \includegraphics[width=1\linewidth]{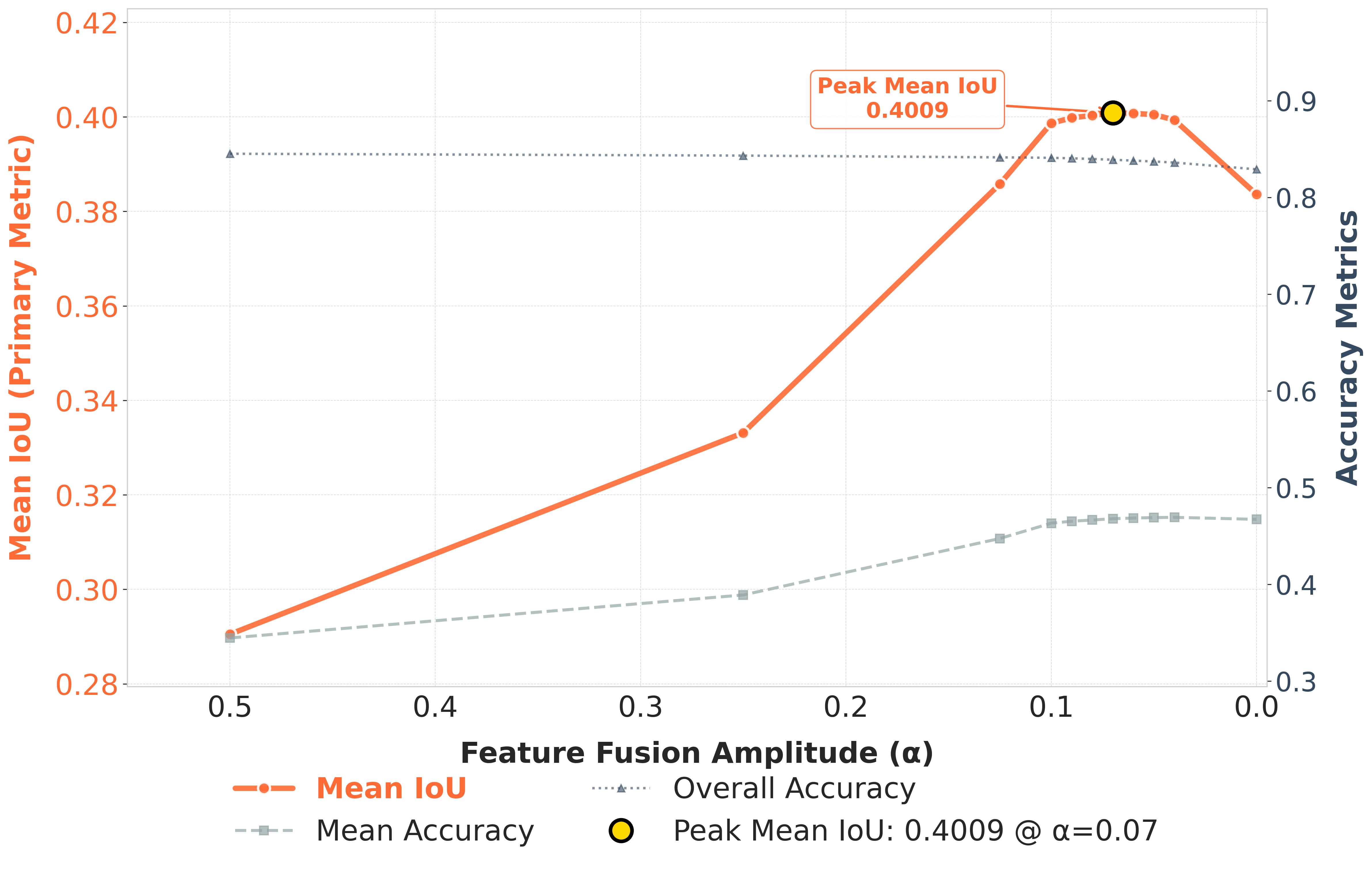}
    \caption{The impact of the weighted fusion coefficient $a$ on model performance on the SensatUrban dataset. }
    \label{fig:sensaturban_fusion}
\end{figure}

\begin{figure}
    \centering
    \includegraphics[width=1\linewidth]{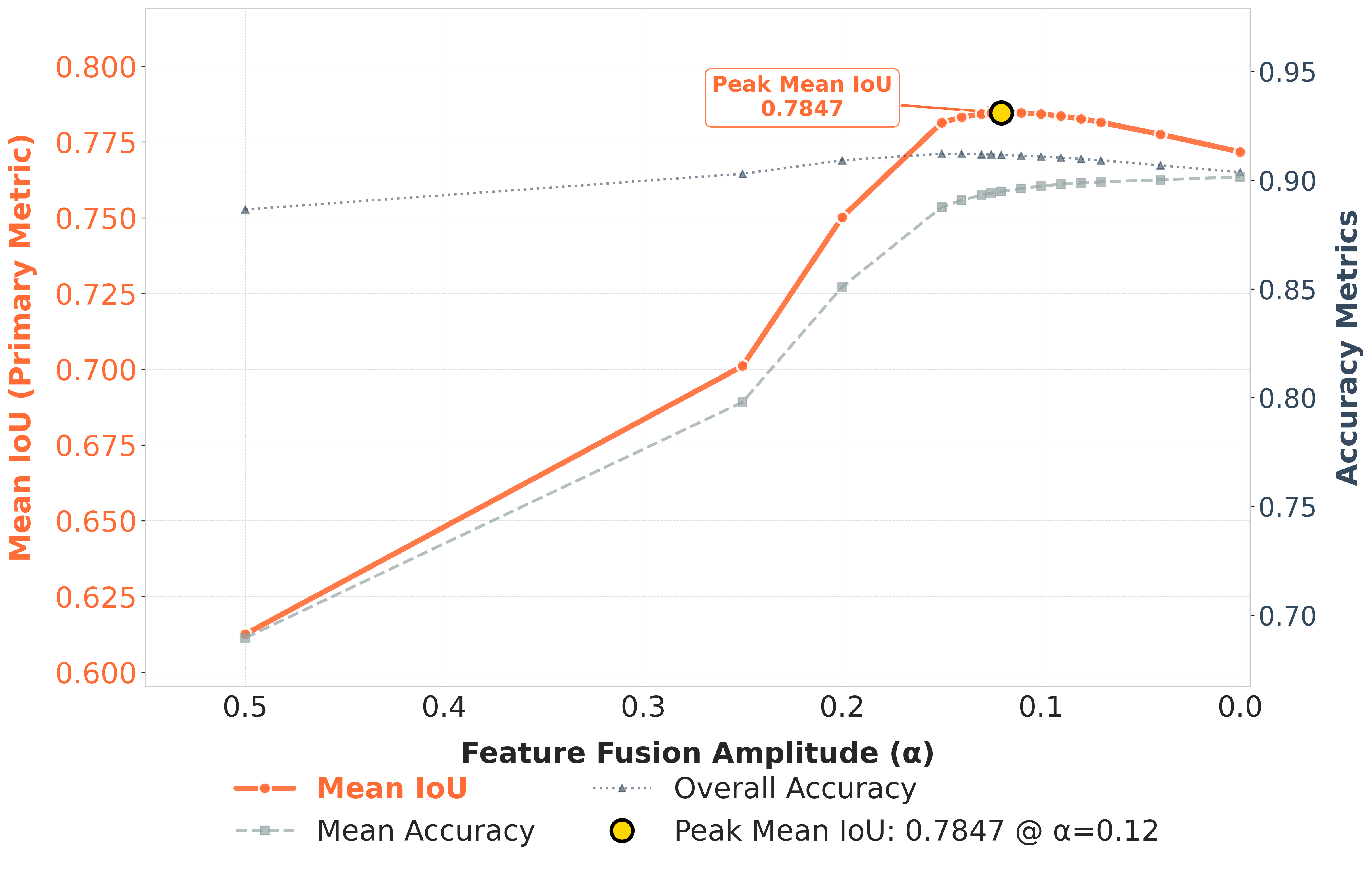}
    \caption{The impact of the weighted fusion coefficient $a$ on model performance on the SUM dataset. }
    \label{fig:fusion_sum}
\end{figure}

\subsubsection{2D-3D Feature Fusion}
To investigate the optimal mixing ratio for our 2D and 3D features during fusion, we conducted a systematic ablation study on the fusion weight, $\alpha$. As illustrated in Figure~\ref{fig:sensaturban_fusion} and Figure~\ref{fig:fusion_sum}, the model's performance exhibits a clear peak effect as the value of $\alpha$ changes. We found that incorporating a small amount of 3D features, controlled by an $\alpha$ value near 0.1, is sufficient to achieve optimal overall performance. Combining this observation with the performance data in Table~\ref{tab:featureablation}, as well as the qualitative examples in Figure~\ref{fig:geometric_edges_sensaturban} and Figure ~\ref{fig:geometric_edges_sum}, we can conclude that the 2D and 3D features play distinct yet complementary roles within our model. The performance curve reaches its apex in the region where $\alpha$ is small (i.e., the weight of 2D features is large), which indicates that rich semantic information is primarily supplied by the 2D features. Therefore, a small amount of 3D features (a small $\alpha$) is sufficient to introduce these critical geometric priors without interfering with the primary semantic recognition, thereby enhancing the model's overall performance.

\begin{figure*}
    \centering
    \includegraphics[height=6cm, keepaspectratio]{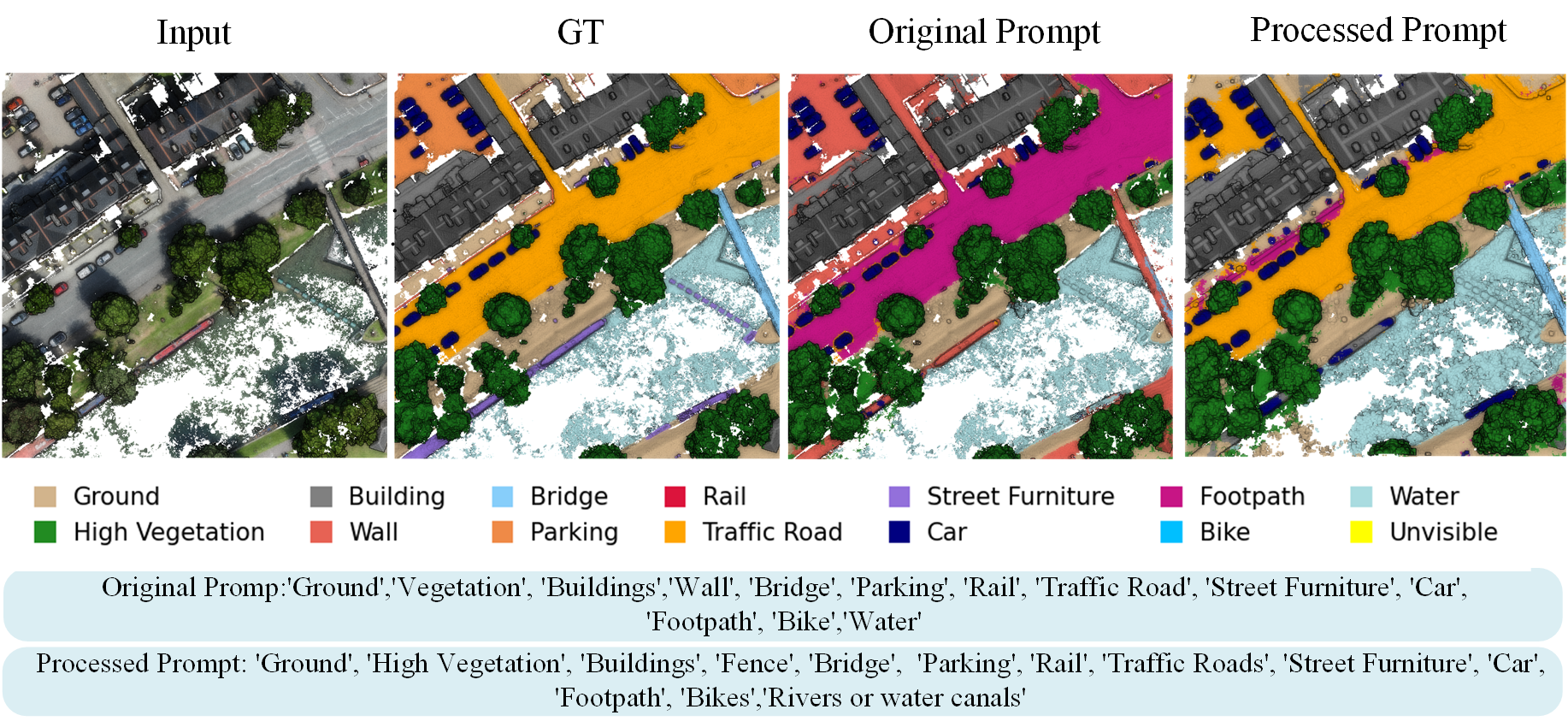}
    \caption{Impact of text prompts on model performance.}
    \label{fig:ab_prompt}
\end{figure*}

\subsubsection{Impact of Prompts on Model Performance}
To investigate the direct impact of text prompt quality on model performance during the inference stage, we conducted a comparative experiment, with the results presented in Table~\ref{tab:ab_sbff_text} and Figure ~\ref{fig:ab_prompt}. The experimental results reveal the critical role of a prompt's semantic precision. When using a relatively ambiguous original prompt, for instance, using ``wall'' to refer to all wall-like structures (including fences and barriers), the model achieved an mIoU of only 30.7\%. We observed that this ambiguous prompt caused the model to incorrectly classify a large number of building facades as the ``wall'' category. This semantic confusion directly led to a severely degraded IoU for the ``building'' class (only 73.1\%). However, when we switched to an optimized prompt with more specific and unambiguous semantics (denoted as ``OpenUrban3D (2D-3D) w/ Processed Prompt''), the model's mIoU significantly increased to 39.6\%. This result provides compelling evidence that in open-vocabulary segmentation tasks, selecting semantically precise and unambiguous prompts is crucial for guiding the model to perform correct semantic reasoning and avoid class confusion.

\section{Limitations and Future Work}
Despite the promising results of OpenUrban3D, several limitations remain, which also point to directions for future research. Due to the massive scale of urban point clouds and GPU memory constraints, we adopt a larger voxel size during the distillation of the 3D backbone, which limits the model's ability to capture fine-grained details. Developing more efficient segmentation algorithms that balance global context and local detail for large-scale point clouds is therefore an important future direction.

Color information plays a significant role in the distillation of 3D features. In large-scale urban datasets, particularly those lacking high-quality imagery such as that provided by RGB-D sensors in indoor environments, uneven illumination and shadows, for example those cast between buildings, can introduce erroneous color cues that negatively affect the learning of the 3D backbone. Future work could explore enhancing 2D features by leveraging only geometric priors from point clouds, thereby reducing reliance on unstable color information.

Finally, the current fusion method for 2D and 3D features is relatively simple. Designing more advanced fusion modules could enable deeper and more effective integration of 2D semantic information and 3D geometric structure, further improving segmentation performance.

\section{Conclusion}
In this paper, we proposed OpenUrban3D, a 3D open-vocabulary semantic segmentation framework tailored for large-scale urban point clouds. Unlike previous methods for indoor or autonomous driving scenes that rely on well-aligned 2D video sequences, OpenUrban3D requires only 3D point clouds as input, greatly reducing data acquisition requirements. The core of our approach lies in generating and distilling robust semantic features without real images. We first render the point cloud from diverse virtual viewpoints using a multi-view, multi-granularity projection module to capture objects at multiple scales, and then apply a mask-based 2D feature extraction module whose outputs are distilled into a 3D backbone. At inference, 2D and 3D features are fused for open-vocabulary segmentation. Experiments show that our method approaches the performance of fully supervised baselines on SensatUrban and even surpasses state-of-the-art fully supervised methods on the SUM dataset. We believe this work lays a solid foundation for advancing open-vocabulary semantic segmentation of large-scale urban point clouds and opens new avenues for robust and scalable 3D scene understanding.

\bibliographystyle{IEEEtran}
\bibliography{ref}


\end{document}